\documentclass[journal]{IEEEtran}
\usepackage{amsmath,amsfonts}
\usepackage{amsthm}
\usepackage{utfsym}
\usepackage{bbm}
\usepackage{algorithm}
\usepackage{algorithmic}
\usepackage{array}
\usepackage[caption=false,font=normalsize,labelfont=sf,textfont=sf]{subfig}
\usepackage{textcomp}
\usepackage{stfloats}
\usepackage{url}
\usepackage{verbatim}
\usepackage{graphicx}
\usepackage{cite}
\usepackage{booktabs}
\usepackage{diagbox}  % 引入diagbox包
\usepackage{array}    % 用于表格居中
\usepackage{makecell}
\usepackage{orcidlink} 
\hypersetup{hidelinks, colorlinks=true, allcolors=blue, pdfstartview=Fit, breaklinks=true}

\newtheorem{definition}{Definition}
\newtheorem{theorem}{Theorem}

\newtheorem{lemma}{Lemma}
\newtheorem{remark}{Remark}
\newtheorem{assumption}{Assumption}
\newtheorem{corollary}{Corollary}

\begin{document}
\title{Analysis of Asynchronous Federated Learning: Unraveling the  Interactions between Gradient Compression, Delay, and Data Heterogeneity}

% <-this % stops a space


\author{Diying Yang, Yingwei Hou, Weigang Wu}
\maketitle

\begin{abstract}

In practical federated learning (FL), the large communication overhead between clients and the server is often a significant bottleneck. Gradient compression methods can effectively reduce this overhead, while error feedback (EF) restores model accuracy. Moreover, due to device heterogeneity, synchronous FL often suffers from stragglers and inefficiency—issues that asynchronous FL effectively alleviates. However, in asynchronous FL settings—which inherently face three major challenges: asynchronous delay, data heterogeneity, and flexible client participation—the complex interactions among these system/statistical constraints and compression/EF mechanisms remain poorly understood theoretically. In this paper, we fill this gap through a comprehensive convergence study that adequately decouples and unravels these complex interactions across various FL frameworks. We first consider a basic asynchronous FL framework AsynFL, and establish an improved convergence analysis that relies on fewer assumptions and yields a superior convergence rate than prior studies. We then extend our study to a compressed version, AsynFLC, and derive sufficient conditions for its convergence, indicating the nonlinear interaction between asynchronous delay and compression rate. Our analysis further demonstrates how asynchronous delay and data heterogeneity jointly exacerbate compression-induced errors, thereby hindering convergence. Furthermore, we study the convergence of AsynFLC-EF, the framework that further integrates EF. We prove that EF can effectively reduce the variance of gradient estimation under the aforementioned challenges, enabling AsynFLC-EF to match the convergence rate of AsynFL. We also show that the impact of asynchronous delay and flexible participation on EF is limited to slowing down the higher-order convergence term. Experimental results substantiate our analytical findings very well.

%, and such an impact is exacerbated by high data heterogeneity
%we fill this gap by analyzing the convergence behaviors of FL under different frameworks.
%Then, we consider a variant framework with gradient compression, AsynFLC
%There is a significant lack of systematic convergence analysis that adequately captures these complex couplings.

\end{abstract}


\section{Introduction}
\IEEEPARstart{F}{ederated} learning (FL) \cite{DBLP:conf/aistats/McMahanMRHA17} is a popular large-scale machine learning paradigm, where a large number of resource-constrained client devices, such as smartphones, personal computers, and edge devices,  collaborate to learn a global model through communication with a server. These clients keep their private data locally and optimize local models by performing multiple SGD steps in one global round. The server aggregates model updates (or gradients) from clients and produces a new global model. 

Since model updates are exchanged between clients and the server in each round, the large communication overhead has always been a major challenge of FL. Gradient compression is an effective technique for reducing communication costs in distributed SGD and FL. Unbiased compressors such as QSGD \cite{DBLP:conf/nips/AlistarhG0TV17} and Stochastic Quantization \cite{DBLP:conf/icml/SureshYKM17} can achieve good convergence performance but low compression ratios. In contrast, biased gradient compression including Top$_k$ sparsification \cite{DBLP:conf/nips/StichCJ18}, SignSGD \cite{DBLP:conf/icml/KarimireddyRSJ19} and so on, can achieve high compression ratios but also introduce compression errors, which will affect convergence. By applying error feedback (EF) \cite{DBLP:conf/interspeech/SeideFDLY14}, using biased compression in FL can achieve nearly the same convergence rate as the full-precision counterpart.

Biased compression and EF are primarily developed and theoretically analyzed for the synchronous aggregation scheme \cite{DBLP:conf/iclr/YangFL21,DBLP:conf/icml/WangLC22,DBLP:conf/icml/ChoSJ0KZ23,DBLP:conf/iclr/HuangLL24}, but their impact on the convergence behaviors of asynchronous FL still lacks comprehensive investigation. Asynchronous FL enables clients to update their models/gradients asynchronously without waiting for the slower ones, thereby effectively alleviating the straggler and inefficiency issues caused by the system-level challenge of device heterogeneity. However, several key features of asynchronous FL, including asynchronous delay, data heterogeneity, and flexible participation, pose challenges when integrating biased compression and EF. Asynchronous delay is caused by the asynchronous aggregation of model updates from clients with different model versions. Data heterogeneity refers to the property that, training data locally kept by clients is non-independent and identically distributed (Non-IID). Flexible participation indicates variability in client participation, meaning the changes of the client set from round to round, with  non-uniform distributions of participation probabilities among clients. These three challenging features interact intricately with gradient compression and EF, complicating the analysis. There remains a significant lack of systematic convergence analysis that adequately unravels these complex interactions.

%\IEEEpubidadjcol

In this paper, we address these challenges, and conduct a systematic study on how biased compression, EF mechanisms, and three inherent challenges—asynchronous delay, data heterogeneity, and flexible participation—collectively influence the convergence behaviors of asynchronous FL under various frameworks. Our investigation proceeds gradually, unraveling and decoupling their intricate interactions. More precisely, we make the following major contributions:

(1) We first provide an improved convergence analysis for a basic asynchronous FL framework, AsynFL. Our analysis depends on fewer assumptions and achieves a better convergence rate than previous work. We do not require the assumption of uniform participation, i.e., clients may participate in global update rounds with non-uniform probabilities. Such flexible participation is more reasonable and practical, despite complicating convergence analysis. We prove that AsynFL achieves a convergence rate of $\mathcal{O}(\frac{1}{\sqrt{TKn}})$ w.r.t. the total communication rounds $T$, the local iterations $K$ and the number of clients $n$. Our analysis also provides insights into the interplay between asynchronous delay and data heterogeneity. 

(2) We then, for the first time, conduct a convergence analysis for asynchronous FL with biased compression, AsynFLC. Removing the assumption of bounded gradients and carefully bounding the compression errors under asynchronous updates and flexible participation, we derive sufficient conditions for convergence that describe the interaction between asynchronous delay and compression rate. Our analysis also demonstrates that gradient compression under asynchronous delay causes larger variance that hampers convergence, and such an impact is exacerbated by high data heterogeneity.

(3) Furthermore, we study the convergence behaviors of AsynFLC-EF, the framework that integrates both biased compression and EF. By simultaneously considering all three challenging features discussed above and addressing their interactions with compression and EF, we prove that EF can effectively mitigate the variance of gradient estimation. This enables AsynFLC-EF to achieve a similar convergence rate as AsynFL. We further demonstrate that the impact of asynchronous delay and flexible participation on EF is limited to a slowdown in the higher-order convergence term. These findings indicate that AsynFLC-EF is robust against heterogeneous, compressed, and delayed gradients.

(4) Finally, we conduct extensive experiments and the results substantiate our analytical findings. It is demonstrated that biased gradient compression makes AsynFLC difficult to converge, while AsynFLC-EF restores the same convergence rate as AsynFL.

\section{Related Work}

\subsection{Convergence Analysis of asynchronous FL}

Almost all studies on the convergence of asynchronous FL \cite{DBLP:conf/aistats/NguyenMZYR0H22,DBLP:conf/allerton/ToghaniU22,DBLP:journals/jmlr/FraboniVKL23,DBLP:journals/tmc/WuBD24,DBLP:conf/iclr/WangCWCC24,DBLP:conf/icml/WangWLC24} do not consider biased gradient compression, and rely on a rather demanding assumption that clients participate with uniform probability. Moreover, Nguyen et al. \cite{DBLP:conf/aistats/NguyenMZYR0H22} fix the number of participating clients in each round. Wang et al. incorporate asynchronous training into adaptive federated optimization \cite{DBLP:conf/icml/WangWLC24} and also mitigate the dependency on the maximum delay $\tau_{max} $ in the convergence for FedBuff \cite{DBLP:conf/iclr/WangCWCC24}.
%Their analyses of asynchronous FL depend on the demanding assumption that clients participate in global aggregation with an equal probability. 
Wang et al. \cite{DBLP:journals/tmc/WangLJZ25} samples $m$ clients uniformly with replacement to ensure linear speedup. Li et al. \cite{DBLP:journals/pami/LiYRSZ24} also assume clients uniformly participate when analyzing the convergence of asynchronous FL with DP. In contrast, our analysis does not require the assumption of uniform participation, and enables clients to participate in the global update with non-uniform probabilities. This is more reasonable and practical, although it makes the convergence analysis more complicated.

Even et al. \cite{DBLP:conf/aistats/EvenKM24}, Bornstein et al. \cite{DBLP:conf/iclr/BornsteinRWBH23} study the convergence of asynchronous updates in decentralized FL. Fraboni et al. \cite{DBLP:journals/jmlr/FraboniVKL23} introduce stochastic aggregation weights to represent the variability of clients' update times. Iakovidou et al. \cite{DBLP:journals/corr/abs-2405-10123} correct the client drift caused
by heterogeneous client update frequencies. Their analyses both focus on convex optimization, while our analysis focuses on non-convex optimization.

%Their analyses \yrcite{DBLP:journals/jmlr/FraboniVKL23,DBLP:journals/corr/abs-2405-10123} focus on convex optimization, while our analysis focuses on non-convex optimization.

\subsection{Gradient Compression and Analysis}

%Gradient compression is an effective technique to reduce communication costs in distributed SGD and FL. 
Various gradient compression methods have been proposed for distributed learning including FL, and 
they can be categorized into unbiased compression and biased compression. Unbiased compressors mainly include randomized quantizers, such as QSGD \cite{DBLP:conf/nips/AlistarhG0TV17} and Stochastic Quantization \cite{DBLP:conf/icml/SureshYKM17}. Quantization refers to reducing the representation precision of each element value in the gradients. QSGD \cite{DBLP:conf/nips/AlistarhG0TV17} can quantize gradients into different levels, such as 2, 4, or 8 bits. Many gradient compression methods using unbiased quantizers have been proposed and analyzed in \cite{DBLP:conf/nips/AlistarhG0TV17,DBLP:conf/nips/WenXYWWCL17,DBLP:journals/tnn/XuDJHC22,DBLP:journals/tifs/LyuHRGYCT24,DBLP:journals/iotj/ChenLWPS24}. These methods require the quantizers to be unbiased to ensure convergence and do not need EF. 

%The effect of compression achieved by quantification is relatively limited. In contrast, sparsification usually achieves higher compression rates while ensuring convergence and model accuracy. 

Biased compressors mainly include Top$_k$ sparsification\cite{DBLP:conf/nips/StichCJ18,DBLP:conf/nips/AlistarhH0KKR18}, deterministic Sign Quantizer\cite{DBLP:conf/icml/KarimireddyRSJ19}, which produce a biased estimator of the true gradient, consequently introducing biased errors that impair convergence\cite{DBLP:conf/nips/AlistarhH0KKR18,DBLP:conf/aaai/GaoXH21,DBLP:conf/iclr/LiKL22,DBLP:conf/icml/Li023}. Top$_k$, the most popular sparsification technique, selects and uploads only $k$ gradient elements with the largest absolute values. Sign Quantizer retains only the sign (1-bit) information of gradients.
%reduce the variance of compressed gradients and

For convex optimization, convex counter-examples including using Top$_k$ and Sign compressors have been provided to show the non-convergence issue of directly applying biased compression in distributed learning\cite{DBLP:conf/icml/KarimireddyRSJ19,DBLP:journals/jmlr/BeznosikovHRS23}. Li et al.\cite{DBLP:conf/icml/Li023} provide the upper bound for convergence when directly applying biased compression in non-convex synchronous FL and show the non-convergence, which is not applicable to asynchronous FL settings.%Li et al. \cite{DBLP:conf/icml/Li023} provide the upper bound for convergence when directly applying biased compression in non-convex synchronous FL and show the non-convergence, which is not applicable to asynchronous FL settings.  %The upper bound may be too loose. Counter-examples or the lower bound that does not converge to zero, are required to prove the non-convergence issue. 
%It has been shown that directly adopting biased compression in distributed SGD may lead to divergence, through counter examples \cite{DBLP:conf/icml/KarimireddyRSJ19,DBLP:journals/jmlr/BeznosikovHRS23}. 

\subsection{EF and Analysis}

To stabilize convergence, EF is usually used to compensate for compressed gradients.
\cite{DBLP:conf/interspeech/SeideFDLY14,DBLP:conf/nips/0001DKD19,DBLP:conf/iclr/LiKL22,DBLP:conf/icml/WangLC22,DBLP:conf/aaai/GaoXH21,DBLP:conf/icml/Li023,DBLP:conf/icml/SunW0W23}. EF retains the difference between the compressed gradient and the true gradient as compression error, which will be compressed together with the model update of the next participation. It has been proved that, when applying EF, synchronous FL with biased compression can match the convergence rate of the full-precision counterpart \cite{DBLP:conf/icml/Li023}. 

Richt{'{a}}rik et al. \cite{DBLP:conf/nips/RichtarikSF21} propose "EF21" as an alternative to the standard EF. Gruntkowska et al. \cite{DBLP:conf/icml/GruntkowskaTR23} and Zhou et al. \cite{DBLP:journals/tnn/ZhouCC25} analyze EF21. Different from their analyses, our work focuses on the standard EF algorithm (a distinct approach from EF21) and establishes an analytical framework that better aligns with real-world complexities, incorporating: 1) stochastic gradient, 2) local steps, 3) asynchronous updates, 4) data heterogeneity, and 5) partial participation \cite{DBLP:journals/pami/ZhangLDZZX25,DBLP:journals/pami/SunSSDT23}.

Gradient compression methods also become increasingly popular in asynchronous FL, while most studies focus on unbiased quantization. Liu et al. \cite{DBLP:journals/iotj/LiuHYHS24} introduce unbiased quantization into the asynchronous FL. Their analysis %in \cite{DBLP:journals/iotj/LiuHYHS24} 
assumes functions are convex and does not consider data heterogeneity. Bian et al. \cite{DBLP:journals/tvt/BianX24} introduce quantization in asynchronous FL, and their analysis depends on a strong assumption that stochastic gradients are bounded and does not consider heterogeneous updates. Xu et al. \cite{DBLP:journals/tdsc/XuGDGZLZZ24} integrate a blockchain-based semi-asynchronous aggregation scheme with SignSGD, but do not conduct a convergence analysis. Different from these methods, we consider biased gradient compression and EF in asynchronous FL, and carefully estimate heterogeneous updates and the compression errors without the assumption of bounded gradients. Our analysis also considers data heterogeneity and flexible participation. 

\section{Asynchronous FL Framework}

Generally, FL aims to solve an optimization problem:
\begin{equation}
\begin{aligned}
\mathop{\min}_{\mathbf{x} \in \mathbb{R}^d} f\big(\mathbf{x}\big) : =  \frac{1}{n} \sum_{i=1}^n f_i \big(\mathbf{x}\big) ,
\end{aligned}
\end{equation}
where $\mathbf{x}$ represents the global model parameter, and for any client $ i\in[n]$ with a local data distribution $\mathcal{D}_i $, the local loss function is $f_i \big(\mathbf{x}\big) =  \mathbb{E}_{\xi_i \sim \mathcal{D}_i} \big[F_i\big(\mathbf{x} ; \xi_i\big)\big]$. We focus on non-convex optimization, i.e., each local function $ f_i$ is non-convex, which is more general and complex. Particularly, the data distributions $\mathcal{D}_i $ are non-IID, indicating that local functions $ f_i$ differ from each other.

In this paper, we conduct the convergence analysis gradually by considering three successive frameworks.

\subsection{AsynFL}

We first introduce AsynFL, a basic asynchronous federated learning framework, which is a generalization of typical asynchronous FL procedures \cite{DBLP:conf/nips/0001DKD19,DBLP:conf/aistats/NguyenMZYR0H22,DBLP:journals/jmlr/FraboniVKL23,DBLP:journals/tmc/WuBD24}.

In AsynFL, clients start and complete local training asynchronously, participating in global updates at their own pace. Upon receiving local updates from clients, the server generates a new global model and returns it to the participating clients. As a result, clients usually perform local training on different versions of global models, while the server 
asynchronously aggregates local updates with varying delays. Different from the popular framework FedBuff \cite{DBLP:conf/aistats/NguyenMZYR0H22}, AsynFL allows the number of clients participating in each global update to vary flexibly, enabling dynamic participation patterns.

\begin{definition}
\label{def:dex}
(Flexible Participation) For any client $i\in[n]$, let $\mathcal{I}_T^{(i)} = \big \{ t_1^{(i)},t_2^{(i)}, \dots,t_{j_i}^{(i)} \big \} \subseteq [T] $ represent the set of global update rounds in which $i$ participates, where $t_{q}^{(i)} $ is a random variable and $t_{q}^{(i)} < t_{q+1}^{(i)} $ for $q=1,\dots,j_i-1$.
\end{definition}

Specifically, client $i$ performs local training upon receiving the global model $\mathbf{x}_{t_{q}^{(i)}}$, but only contributes its local updates to the global model in the communication round $t_{q+1}^{(i)} $. This indicates a delay of $t_{q+1}^{(i)} -t_{q}^{(i)}$, which may vary for different $q$. And $\mathcal{I}_T^{(i)}$ is not identical for different clients.

\begin{definition}
\label{def:delay} (Asynchronous Delay). Define the random variable $\tau_t^{i} \in [T] $ to represent the delay for any client $i\in[n]$, denoting the difference between the current global round $t$ and the last round where $i$ participated in the global update. 
\end{definition}
%The asynchronous delay $\tau_t^{i}$ depends on network conditions and availability. 

The specific operations of AsynFL are defined as follows. Each client performs $K$ steps of local SGD using local data:
\begin{equation}
\begin{aligned}
\mathbf{x}_{t, k+1}^{(i)}=\mathbf{x}_{t, k}^{(i)}-\eta \nabla F_i\big(\mathbf{x}_{t, k}^{(i)} ; \xi_{t, k}^{(i)}\big).
\end{aligned}
\end{equation}
After $K$ steps, client $i$ obtains the local model $\mathbf{x}_{t, K}^{(i)}$. To calculate the local update $\mathbf{\Delta}_t^{(i)} $, client $i$ computes the difference between the local model $\mathbf{x}_{t, K}^{(i)}$ and the global model $\mathbf{x}_{t-\tau_t^{i}}$, where $t-\tau_t^{i}$ represents the communication round when $i$ began to compute local gradients. 
\begin{equation}
\begin{aligned}
\label{equ:3}
\mathbf{\Delta}_t^{(i)}=\mathbf{x}_{t, K}^{(i)}-\mathbf{x}_{t-\tau_t^{i}}.
\end{aligned}
\end{equation}
When participating in the global update, i.e., $t+1 \in \mathcal{I}_T^{(i)} $, client $i$ uploads its local update $\mathbf{\Delta}_t^{(i)} $ and downloads the updated global model $\mathbf{x}_{t+1} $ that incorporates its contribution. The local model is updated as:
\begin{equation}
\begin{aligned}
{\mathbf{x}_{t+1}^{(i)}} = \begin{cases}
\mathbf{x}_{t}^{(i)}-\eta \sum_{k=0}^{K-1} \nabla F_i(\mathbf{x}_{t, k}^{(i)} ; \xi_{t, k}^{(i)}),&{\text{if}}\ {t+1 \notin \mathcal{I}_T^{(i)}} \\ 
{\mathbf{x}_{t+1},}&{\text{if}}\ {t+1 \in \mathcal{I}_T^{(i)}} 
\end{cases}
\end{aligned}
\end{equation}
\noindent
where the global model $\mathbf{x}_{t+1} $ is acquired by aggregating the local updates from clients in $S_t = \{i|t+1 \in \mathcal{I}_T^{(i)}\} $ as follows:
\begin{equation}
\begin{aligned}
\mathbf{x}_{t+1}=\mathbf{x}_t+  \frac{\eta_g}{n} \sum_{i \in S_t} \mathbf{\Delta}_t^{(i)}.
\end{aligned}
\end{equation}

In AsynFL, no assumptions are made on the number of participating clients $|S_t|$, so it can vary from round to round. This design enables the system to adapt more flexibly to different practical application scenarios, where network conditions or clients' participation may vary from time to time.

\subsection{AsynFLC}

The second framework AsynFLC integrates biased gradient compression with AsynFL.

\begin{definition}
\label{def:com}
(Biased Compressor). A compression operator $ \mathcal{C}$ : $ \mathbb{R}^d \to \mathbb{R}^d $ is a $\gamma-$contraction operator \cite{DBLP:conf/nips/StichCJ18} if there exists a constant $ \gamma \in (0,1] $ such that
\begin{equation}
\begin{aligned}
\mathbb{E}_\mathcal{C}\|\mathbf{x} - \mathcal{C} (\mathbf{x})\|^2 \leq (1-\gamma) \|\mathbf{x}\|^2 , \forall \mathbf{x} \in \mathbb{R}^d.
\end{aligned}
\end{equation}
\end{definition}

If $\gamma = 1$, we have $\mathcal{C} (\mathbf{x}) = \mathbf{x} $, which means $\mathbf{x} $ is not compressed. A smaller $\gamma$ implies a larger degree of compression. 

In our analysis, we consider several representative biased compressors. (1) Top$_k$ sparsification. For any $\mathbf{x} \in \mathbb{R}^d $, Top$_k$$(\mathbf{x}) $ with $\gamma=k/d$ \cite{DBLP:conf/nips/AlistarhH0KKR18}, has at most $k$ non-zero components with the largest absolute value in the $d-$length vector. (2) Deterministic Sign Quantizer \cite{DBLP:conf/icml/BernsteinWAA18,DBLP:conf/icml/KarimireddyRSJ19}. For any $\mathbf{x} \in \mathbb{R}^d $, $i \in [d]$, the $i$'th component of Sign($\mathbf{x}$) is $\mathbbm{1}\{x_i \ge 0\} -\mathbbm{1}\{x_i < 0\} $. (3) The composition of sparsification and quantization \cite{DBLP:conf/nips/0001DKD19}. Combining Top$_k$ and the quantizer $Q_s$ in QSGD \cite{DBLP:conf/nips/AlistarhG0TV17} to achieve higher compression rates, we obtain a biased compressor $Q_s(Top_k(\mathbf{x}))$ with $\gamma = \frac{k}{d(1+\beta_{k,s})}$ , $\beta_{k,s} = \min(\frac{k}{s^2},\frac{\sqrt{k}}{s})$ \cite{DBLP:conf/nips/0001DKD19}.

Using the compressor, clients directly compress the local update $\mathbf{\Delta}_t^{(i)} $ obtained by \eqref{equ:3} in AsynFL, and send the compressed update $\mathcal{C}\big(\mathbf{\Delta}_t^{(i)}\big)$ to the server. Then, the server aggregates $\mathcal{C}\big(\mathbf{\Delta}_t^{(i)}\big)$ from clients $i\in S_t$ and updates the global model as follows:
\begin{equation}
\begin{aligned}
\mathbf{x}_{t+1}=\mathbf{x}_t+  \frac{\eta_g}{n} \sum_{i \in S_t} \mathcal{C}\big(\mathbf{\Delta}_t^{(i)}\big).
\end{aligned}
\end{equation}
Except for the compression operation, the other steps in AsynFLC are the same as in AsynFL.

\subsection{AsynFLC-EF}

The third framework is AsynFLC-EF, which integrates AsynFLC with EF. The operations of this framework are presented as Algorithm \ref{alg:alg1}.

Specifically, each client $i$ maintains the local error accumulator $\mathbf{e}^{(i)}$, initialized as $\mathbf{0}$. It is important to note that the local error accumulator $ \mathbf{e}_t^{(i)} :=  \mathbf{e}_{t-\tau_t^{i}}^{(i)}$, where $\tau_t^{i}$ measures the delay since client $i$’s most recent participation. During the delay period, the update of the local error accumulator is halted. Upon client $i$'s engagement in server aggregation, both the local update $\mathbf{\Delta}_t^{(i)}$ and the local error accumulator $ \mathbf{e}_t^{(i)}$ are compressed prior to transmission as follows:
\begin{equation}
\begin{aligned}
\label{equ:8}
\widehat{\mathbf{\Delta}}_t^{(i)}=\mathcal{C}\big(\mathbf{\Delta}_t^{(i)}+\mathbf{e}_{t-\tau_t^{i}}^{(i)}\big),
\end{aligned}
\end{equation}
where $t-\tau_t^{i} \in \mathcal{I}_T^{(i)} $ represents the last round of global update that client $i$ participates in. 
%It is important to note that $ \mathbf{e}_t^{(i)} =  \mathbf{e}_{t-\tau_t^{i}}^{(i)}$, 
%This indicates a delay of $\tau_t^{i}$ rounds. Actually, the error compensation $\mathbf{e}_{t-\tau_t^{i}}^{(i)}$ is postponed to the round $t$ for use. Due to such asynchrony, the local error accumulator is not actually updated every round. %The client only applies the compressor and calculates the local error when participating in the global aggregation. 

\renewcommand{\algorithmicrequire}{\textbf{Initialize:}}  
\renewcommand{\algorithmicensure}%{\textbf{Output:}} 

\begin{algorithm}[H]
\caption{AsynFLC-EF.}\label{alg:alg1}
\begin{algorithmic}[1]
\REQUIRE global model $\mathbf{x}_0$; local model $\mathbf{x}_0^{(i)}=\mathbf{x}_0$, local error accumulator  $\mathbf{e}_0^{(i)}=\mathbf{0}$, the set of participating rounds $\mathcal{I}_T^{(i)}$, for any client $ i \in[n]$; local learning rate $\eta$, global learning rate $\eta_g$.
\FOR{each round $t = 0, \ldots ,T-1$}
\FOR{each client $i \in[n]$ in parallel}
\STATE $\mathbf{x}_{t, 0}^{(i)}=\mathbf{x}_t^{(i)}$
\FOR{$k = 0, \ldots ,K-1$}
\STATE Compute local stochastic gradient $\nabla F_i\big(\mathbf{x}_{t, k}^{(i)} ; \xi_{t, k}^{(i)}\big)$
\STATE $\mathbf{x}_{t, k+1}^{(i)}=\mathbf{x}_{t, k}^{(i)}-\eta \nabla F_i\big(\mathbf{x}_{t, k}^{(i)} ; \xi_{t, k}^{(i)}\big)$
\ENDFOR
\IF{client $i$ will participate in the global update}
\STATE  $\mathcal{I}_T^{(i)} \leftarrow \{t+1\} \cup \mathcal{I}_T^{(i)}$
\STATE  Compute the local update $\mathbf{\Delta}_t^{(i)}=\mathbf{x}_{t, K}^{(i)}-\mathbf{x}_{t-\tau_t^{i}}$
\STATE  Compress the update $\widehat{\mathbf{\Delta}}_t^{(i)}=\mathcal{C}\big(\mathbf{\Delta}_t^{(i)}+\mathbf{e}_{t-\tau_t^{i}}^{(i)}\big)$
\STATE  Send $\widehat{\mathbf{\Delta}}_t^{(i)}$ to the server
\STATE  Update the error $\mathbf{e}_{t+1}^{(i)}=\mathbf{e}_{t-\tau_t^{i}}^{(i)}+\mathbf{\Delta}_t^{(i)}-\widehat{\mathbf{\Delta}}_t^{(i)}$
\STATE  Receive $\mathbf{x}_{t+1}$ from server and set $\mathbf{x}_{t+1}^{(i)}=\mathbf{x}_{t+1}$
\ELSE
\STATE  $\mathbf{x}_{t+1}^{(i)}=\mathbf{x}_{t, K}^{(i)}$,   $\mathbf{e}_{t+1}^{(i)}=\mathbf{e}_{t-\tau_t^{i}}^{(i)}$
\ENDIF
\ENDFOR
\STATE \textbf{Server does:}
\STATE \hspace{0.3cm} Receive $\widehat{\mathbf{\Delta}}_t^{(i)}$ from client $i$, $i\in S_t = \{i|t+1 \in \mathcal{I}_T^{(i)}\}$
%\STATE \hspace{0.3cm} to server
\STATE \hspace{0.3cm} Aggregate local updates $\widehat{\mathbf{\Delta}}_t=\frac{1}{n} \sum_{i \in S_t} \widehat{\mathbf{\Delta}}_t^{(i)}$
\STATE \hspace{0.3cm} Update global model $\mathbf{x}_{t+1}=\mathbf{x}_t+ \eta_g \widehat{\mathbf{\Delta}}_t$
\STATE \hspace{0.3cm} Broadcast $\mathbf{x}_{t+1}$ to the clients in $S_t$
\ENDFOR
\end{algorithmic}
% \label{alg1}
\end{algorithm}

Client $i$ updates its local error accumulator as follows:
\begin{equation}
\begin{aligned}
\mathbf{e}_{t+1}^{(i)}=\mathbf{e}_t^{(i)}+\mathbf{\Delta}_t^{(i)}-\widehat{\mathbf{\Delta}}_t^{(i)},
\end{aligned}
\end{equation}
where $\mathbf{e}_{t+1}^{(i)}$ represents the residual error between the full-precision update and the compressed one. The error $\mathbf{e}_{t+1}^{(i)}$ will be used to compensate for the compressed update when client $i$ participates the next time.

Equation \eqref{equ:8} indicates a delay of $\tau_t^{i}$ rounds. Actually, the error compensation $\mathbf{e}_{t-\tau_t^{i}}^{(i)}$ is postponed to round $t$ for use. Due to such asynchrony, the local error accumulator is not actually updated every round.

The $\widehat{\mathbf{\Delta}}_t^{(i)}$ that contains delayed gradient information from $ \mathbf{e}_t^{(i)}$ will be sent to the server for aggregation. The global model is updated by aggregating the compressed updates from clients $i \in S_t $ as follows:
\begin{equation}
\begin{aligned}
\mathbf{x}_{t+1}=\mathbf{x}_t+  \frac{\eta_g}{n} \sum_{i \in S_t} \widehat{\mathbf{\Delta}}_t^{(i)}.
\end{aligned}
\end{equation}

\begin{table}[hp]
\renewcommand\arraystretch{1.0}
	\centering  % 显示位置为中间
	\caption{Key Notations}  %%表的标题
	\begin{tabular}{p{10pt}l|l} %第一列设置宽度为45pt 全为左对齐 没有分割线
		\hline  % 表格的横线
		%\toprule % 顶部线
		Symbol & & Description  \\%[3pt]只改一行    %%表格第一行标题 % 表格中的内容，用&分开，\\表示下一行
		\hline  % 表格的横线
		%\midrule % 中部线
		    %%表格内容
		%\midrule
		
		$T$,$t$       & & Total number of global rounds, Global round           \\
            $n$       & & Total number of clients           \\
            $K $       & & Number of stochastic gradient descent (SGD) iterations
            \\
            $\eta $,$\eta_g $       & & Local learning rate, Global learning rate
            \\
            $\mathbf{x}_t$   & & Global model parameters at global round $t$            \\

            $\mathbf{x}_{t, k}^{(i)} $       & & Local model parameters of client $i$ updated after the $k$-th SGD
            \\
            $\mathbf{x}_{t}^{(i)} $       & & Local model parameters of client $i$ at global round $t$
            \\
            $\mathbf{\Delta}_t^{(i)} $       & & Local model update of client $i$ at global round $t$
            \\
            $\widehat{\mathbf{\Delta}}_t^{(i)} $       & & Compressed local model update of client $i$ at global round $t$
            \\
            $\mathcal{C}\big(\cdot) $,$\gamma$       & & Compression operator, Compression coefficient
            \\
            $ \mathbf{e}_t^{(i)}$       & & Residual error of client $i$ at global round $t$
            \\
            $S_t $       & & Subset of clients participating at global round $t+1$
            \\
            $t_{q}^{(i)} $       & & The $q$-th global update round in which client $i$ participates       \\
            $\mathcal{I}_T^{(i)}$       & & Set of global rounds in which client $i$ participates within $T$           \\
            $\tau_t^{i} $       & & Asynchronous delay of client $i$ at global round $t$
            \\
            $\tau_{max} $       & & Maximum delay
            \\
            $\tau_{avg_0} $       & & Average delay
            \\
            $\tau_{avg_1} $       & & Average inter-participation delay
            \\
            $\tau_{avg_{m0}} $       & & Average per-round maximum delay
            \\
            $\tau_{avg_{m1}} $       & & Average maximum inter-participation delay
            \\
            $\sigma $       & & Local variance of stochastic gradients (constant)
            \\
            $\sigma_g $       & & Global variance (constant)
            \\
            
		%\bottomrule % 底部线
		\hline  % 表格的横线
	\end{tabular}
\end{table}

\section{Convergence Analysis}

In this section, we analyze the convergence behaviors of all three frameworks for non-convex optimization.

\subsection{Assumptions and Definitions}

\begin{assumption}
(Smoothness). For $\forall i \in[n]$, the local objective function $f_i$ is $L$-smooth: $\forall \mathbf{x}, \mathbf{y} \in \mathbb{R}^d $ , $ \left\|\nabla f_i(\mathbf{x})-\nabla f_i(\mathbf{y})\right\| \le L\|\mathbf{x}-\mathbf{y}\| \text {. } $
\label{ass:1}
\end{assumption}

Assumption \ref{ass:1} is standard in federated optimization \cite{DBLP:journals/tnn/YanTW24,DBLP:conf/aistats/ChenLC24,DBLP:conf/icml/AllouahFGGPRV24}. The Lipschitz gradient condition implies that the global function $f$ is also $L$-smooth and $f(\mathbf{y}) \leq f(\mathbf{x}) + \langle \nabla f(\mathbf{x}), \mathbf{y} - \mathbf{x}\rangle + \frac{L}{2}\|\mathbf{y}-\mathbf{x}\|^2 $ holds for any $ \mathbf{x}, \mathbf{y} \in \mathbb{R}^d $.

\begin{assumption}
(Bounded Variance). $\forall i \in[n]$, $\forall \mathbf{x} \in \mathbb{R}^d$: (i) the stochastic gradient is unbiased: $\mathbb{E}_{\xi \sim D_i} \nabla F_i(\mathbf{x} ; \xi)=\nabla f_i(\mathbf{x})$; (ii) the local variance of the stochastic gradient is bounded: $ \mathbb{E}\left\|\nabla F_i(\mathbf{x} ; \xi)-\nabla f_i(\mathbf{x})\right\|^2 \le \sigma^2$; (iii) the global variance of the gradient is bounded: $ \frac{1}{n} \sum_{i=1}^n\left\|\nabla f_i(\mathbf{x})-\nabla f(\mathbf{x})\right\|^2 \le \sigma_g^2$.
\label{ass:2}
\end{assumption}

Assumption \ref{ass:2} is widely used in the federated setting \cite{DBLP:conf/icml/WangLC22,DBLP:conf/icml/Li023,DBLP:conf/iclr/WangCWCC24}, where the local functions $f_i$ are heterogeneous. The global variance characterizes the data heterogeneity among clients. When clients have identical data distributions, $\sigma_g=0$.

\textbf{It is important to note that, different from existing works \cite{DBLP:conf/aistats/NguyenMZYR0H22,DBLP:conf/allerton/ToghaniU22,DBLP:conf/iclr/WangCWCC24,DBLP:conf/icml/WangWLC24,DBLP:journals/tmc/WangLJZ25,DBLP:journals/pami/LiYRSZ24}, we do not require the assumption that all clients participate with uniform probability. Our analysis also does not assume that the gradients are bounded, while most existing studies make this assumption \cite{DBLP:conf/icml/WangLC22,DBLP:conf/icml/Li023,DBLP:conf/iclr/WangCWCC24,DBLP:journals/tmc/WangLJZ25}.} With fewer assumptions, our convergence analysis is more practical and accurate.

\begin{definition}
\label{def:d}
(Asynchronous Delay). Define the maximum delay as $\tau_{max} = \mathop{\max}_{t \in [T], i \in [n]}  \{ \tau_t^{i}  \}$; the average delay as $\tau_{avg_0} = \frac{1}{T} \sum_{t=0}^{T-1} \frac{1}{n} \sum_{i=1}^n \tau_t^{i} $; the average of the maximum delay over time as $\tau_{avg_{m0}} = \frac{1}{T} \sum_{t=0}^{T-1}   \tau_t^{max} = \frac{1}{T} \sum_{t=0}^{T-1} \mathop{\max}_{i \in [n]}  \{ \tau_t^{i}  \}  $.
\end{definition}

It is common to assume that the maximum delay satisfies $\tau_{max} < \infty $ \cite{DBLP:conf/aistats/NguyenMZYR0H22,DBLP:conf/allerton/ToghaniU22,DBLP:conf/nips/KoloskovaSJ22,DBLP:conf/iclr/WangCWCC24,DBLP:conf/icml/WangWLC24}. To simplify the presentation, we also define another form of the average delay as $\tau_{avg_1} = \frac{1}{T} \sum_{t=0}^{T-1} \frac{1}{n} \sum_{i=1}^n \tau_{t-\tau_t^{i}}^{avg} $ and another form of the average of the maximum delay as $\tau_{avg_{m1}} = \frac{1}{T} \sum_{t=0}^{T-1} \frac{1}{n} \sum_{i=1}^n  \tau_{t-\tau_t^{i}}^{max} $, where $\tau_{t-\tau_t^{i}}^{avg} = \frac{1}{n} \sum_{s=1}^n \tau_{t-\tau_t^{i}}^{s} $ and $\tau_{t-\tau_t^{i}}^{max}=\mathop{\max}_{s \in [n]}  \{ \tau_{t-\tau_t^{i}}^{s}  \} $. The relationships among these forms of delay are as follows: $\tau_{avg_1}<\tau_{avg_0} < \tau_{avg_{m1}} < \tau_{avg_{m0}} \ll \tau_{max}$.

\subsection{Convergence Analysis of AsynFL}

Here, we analyze the convergence in the non-convex case for the full-precision asynchronous FL framework and present the results. The proofs are provided in Appendix A of the supplementary material.

\begin{theorem}
\label{thm:1}
(Convergence of AsynFL). Suppose Assumption \ref{ass:1} and Assumption \ref{ass:2} hold. If the local learning rates satisfy $\eta \leq \frac{1}{36 \sqrt{2} \tau_{max}^{1.5} \eta_g K L} $, AsynFL satisfies:
\begin{equation}
\begin{aligned}
\label{ine:11}
& \frac{1}{T}  \sum_{t=0}^{T-1}  \mathbb{E}\|\nabla f (\mathbf{x}_t)\|^2 \leq  \frac{8[ f(\mathbf{x}_0) - f (\mathbf{x}^{*}) ]}{\eta \eta_g K T}  +  \frac{4 \eta \eta_g  L\sigma^2}{n}  + \\ &  \eta^2 K L^2 \big[4(\sigma^2 + 6 K \sigma_g^2)  + \frac{1}{n} ( 33 \tau_{avg_0} + 72 \tau_{avg_1} )  \eta_g^2  \sigma^2 \big] \\ & + ( \lambda_1 + \lambda_2 )\eta^2  K L^2 \sigma^2    +  (\varphi_1 + \varphi_2 +\varphi_3 )\eta^2  K^2 L^2 \sigma_g^2 ,
\end{aligned}
\end{equation}
where $\mathbf{x}^{*}=arg \min f(\mathbf{x})$, $ \lambda_1= 72\tau_{avg_0}  + 144  \tau_{avg_0} \tau_{max}  \eta^2  K^2 L^2$, $\lambda_2=( 18  \tau_{avg_0} \tau_{max}  +  144 \tau_{max} \tau_{avg_1} ) \eta^2 \eta_g^2 K^2 L^2 $, $\varphi_1 = \tau_{max}  \tau_{avg_{m0}}  ( 288  + 1728 \eta^2  K^2 L^2)$, $\varphi_2 = \tau_{max}  \tau_{avg_{m1}}  ( 288   \eta_g^2  + 1728 \eta^2 \eta_g^2 K^2 L^2)$, $\varphi_2 = \tau_{max}  \tau_{avg_{m0}}  ( 36   \eta_g^2  + 216 \eta^2 \eta_g^2 K^2 L^2)$.
\end{theorem}

%We first describe some important terms in the upper bound and elaborate on their origin in Equation \eqref{ine:11} . The upper bound is comprised of the optimization part (the first term) and the error part (the remaining terms). 

%The optimization part depends on the initialization through the distance between the initial model $\mathbf{x}_0$ and the optimal $\mathbf{x}^*$ on the objective function $f$, which is standard for SGD optimization. It can be minimized by performing sufficient local SGD steps $KT$ and choosing a preferable global learning rate $\eta_g$. 

The upper bound is comprised of the optimization part (the first term) and the error part (the remaining terms). The optimization part depends on the initialization, which is standard for SGD optimization. The error part involves the local stochastic variance $\sigma$, global variance $\sigma_g$ and gradient delay including $\tau_{max}$, $\tau_{avg}$ and so on. The local stochastic variance $\sigma$ is caused by stochastic gradient descent. The term containing $\sigma_g$ is proportional to the data heterogeneity of clients, which accounts for the differences in clients' updates. Asynchronous delay reflects gradient staleness, causing an increase in the error part compared to the synchronous case.

1) Eliminating $\sigma$ (or setting $\sigma=0$), AsynFL reduces to full gradient descent.

2) Eliminating $\sigma_g$ (or setting $\sigma_g=0$) and keeping one local step $K=1$, AsynFL reduces to distributed learning with IID data.

3) Eliminating asynchronous delay, i.e., setting $\tau_{max}=\tau_{avg}=1$, AsynFL reduces to synchronous FL. 

The upper bound provided by Theorem \ref{thm:1} will converge to zero as $T$ increases, indicating that AsynFL converges to a first-order stationary point. %When the local learning rate $\eta$ decreases, the error part also decreases, whereas the optimization part is simultaneously affected by $\eta$ and $T$. 
To ensure optimal convergence, it is essential to select appropriate learning rates.

%The convergence bound on the norm-squared of the gradient, provided by Theorem 1, will converge to zero as $T$ increases, indicating that the algorithm converges to a first-order stationary point. When the local learning rate $\eta$ decreases as $T$ increases, the error part is decreasing, whereas the optimization part is simultaneously affected by the decreasing learning rate $\eta$ and the increasing $T$. To ensure that the algorithm converges to the optimum in asynchronous federated optimization, it is essential to select an appropriate local learning rate.

%In the following, we will present and discuss Corollary 1 obtained by selecting some appropriate learning rates.

\begin{corollary}
\label{cor:1}
Suppose the conditions in Theorem \ref{thm:1} are satisfied. Let $\triangle =f(\mathbf{x}_0) - f (\mathbf{x}^{*})$. If choosing the learning rates  $\eta=\Theta\big(\frac{1}{K \sqrt{T}}\big)$, $\eta_g=\Theta\big( \sqrt{Kn}\big)$, AsynFL satisfies:
\begin{equation}
\begin{aligned}
\frac{1}{T} & \sum_{t=0}^{T-1}  \mathbb{E}\|\nabla f(\mathbf{x}_t)\|^2 = \mathcal{O}\Big(\frac{\triangle}{\sqrt{TKn}}+\frac{\sigma^2}{\sqrt{TKn}} \\ & +\frac{\sigma^2+K \sigma_g^2 }{TK}  +\frac{ \tau_{avg} \sigma^2 + Kn\tau_{max}\tau_{avg_m}  \sigma_g^2}{T} \Big) , 
\end{aligned}
\end{equation}
where $\tau_{avg} = \max \{\tau_{avg_0}, \tau_{avg_1} \}$, $\tau_{avg_m} = \max \{\tau_{avg_{m0}}, \tau_{avg_{m1}}\}$. 
\end{corollary}

\begin{remark}
Corollary \ref{cor:1} suggests that AsynFL achieves a desired convergence rate of $\mathcal{O}\big(\frac{1}{\sqrt{TKn}}\big)$ for a sufficiently large $T$, where $T$ is the number of communication rounds, $K$ is the number of local steps, and $n$ is the number of clients. To reach a $\epsilon-$stationary point, i.e., $\frac{1}{T}  \sum_{t=0}^{T-1} \mathbb{E}\|\nabla f(\mathbf{x}_t)\|^2 \leq \epsilon$, we obtain a communication round complexity of $\mathcal{O}\big(\frac{1}{Kn\epsilon^2}\big)$. This indicates that, when $T$ is sufficiently large and the maximum delay $\tau_{max}$ is relatively small, AsynFL can achieve a linear speedup and match the convergence rate of non-convex synchronous FL on non-IID data \cite{DBLP:conf/iclr/YangFL21,DBLP:conf/iclr/ReddiCZGRKKM21,DBLP:conf/icml/WangLC22,DBLP:journals/pami/SunSSDT23}. 
\end{remark}

\begin{table*}
\centering
\renewcommand\arraystretch{1.0}
	\centering  % 显示位置为中间
	\caption{Comparison of our analysis with asynchronous FL methods under the same learning rate}  %%表的标题
	\begin{tabular*}{\linewidth}{m{2.6cm}|m{0.6cm}<{\centering}|m{1.38cm}<{\centering}|m{1.5cm}<{\centering}|m{0.5cm}<{\centering}|m{1.3cm}<{\centering}|m{1.6cm}
    <{\centering}|l<{\centering}}
		\hline  % 表格的横线
		%\toprule % 顶部线
		 \textbf{Related Work}  &  \textbf{Non-IID} & \textbf{Quantization} & \textbf{Sparsification} & \textbf{EF} & \textbf{Unbounded Gradient} & \textbf{Non-uniform participation} %& \textbf{Asyn} 
         & \textbf{Convergence Rate}    \\%[3pt]只改一行    %%表格第一行标题 % 表格中的内容，用&分开，\\表示下一行
		\hline  % 表格的横线
		%\midrule % 中部线
		    %%表格内容
		%\midrule
        %%ICML'23\cite{DBLP:conf/icml/Li023}       & \usym{1F5F8}  & \usym{1F5F8} &  \usym{1F5F8} & \usym{1F5F8}  & \usym{1F5F8}  & \usym{2715} &  \usym{2715} & $\mathcal{O}\big(\frac{\sigma^2}{\sqrt{TKm}}\big)+\mathcal{O}\big(\frac{\sqrt{K}\sigma_g^2}{\sqrt{Tm}}\big)  $$+\mathcal{O}\big(\frac{ \tau_{max}\tau_{avg_m} }{T}\big)$         \\ \hline
		AISTATS'22\cite{DBLP:conf/aistats/NguyenMZYR0H22}  & \usym{1F5F8}  & \usym{2715} & \usym{2715} & \usym{2715}  & \usym{2715} & \usym{2715} %&  \usym{1F5F8}
        & $\mathcal{O}\big(\frac{\sigma^2}{\sqrt{TK}}\big)$$+\mathcal{O}\big(\frac{\tau_{max}^2 (\sigma^2+\sigma_g^2+G^2)}{T}\big)$       \\ \hline
        Allerton'22\cite{DBLP:conf/allerton/ToghaniU22}  & \usym{1F5F8}  & \usym{2715} & \usym{2715} & \usym{2715}  & \usym{1F5F8} & \usym{2715} %& \usym{1F5F8} 
        & $\mathcal{O}\big(\frac{\sigma^2}{\sqrt{T}}\big)$$+\mathcal{O}\big(\frac{ \sigma_g^2}{\sqrt{T}}\big)$$+\mathcal{O}\big(\frac{\tau_{max}^2 (\sigma^2+\sigma_g^2)}{T}\big)$       \\ \hline
        TPAMI'24\cite{DBLP:journals/pami/LiYRSZ24}       & \usym{1F5F8}  & \usym{2715} & \usym{2715} & \usym{2715}  & \usym{2715} & \usym{2715} %& \usym{1F5F8} 
        & $\mathcal{O}\big(\frac{\sigma^2}{\sqrt{T}}\big)$$+\mathcal{O}\big(\frac{ \sigma_g^2}{\sqrt{T}}\big)$$+\mathcal{O}\big(\frac{\tau_{c}^2 (\sigma^2+G^2)}{T}\big)$
        \\ \hline
        
        TVT'24\cite{DBLP:journals/tvt/BianX24}       & \usym{2715}  & \usym{1F5F8} & \usym{2715} & \usym{2715}  & \usym{2715} & \usym{2715} %&  \usym{1F5F8} 
        & $\mathcal{O}\big(\frac{\sigma^2}{\sqrt{Tn}}\big)$$+\mathcal{O}\big(\frac{\tau_{max}^2 G^2}{T}\big)$
        \\ \hline
        ICLR'24\cite{DBLP:conf/iclr/WangCWCC24}       & \usym{1F5F8}  & \usym{2715} & \usym{2715} & \usym{2715}  & \usym{1F5F8} & \usym{2715} %&  \usym{1F5F8} 
        & $\mathcal{O}\big(\frac{\sigma^2}{\sqrt{TKm}}\big)+\mathcal{O}\big(\frac{\sqrt{K}\sigma_g^2}{\sqrt{Tm}}\big)  $$+\mathcal{O}\big(\frac{ \tau_{max}\tau_{avg_m} }{T}\big)$
            \\ \hline
        ICML'24\cite{DBLP:conf/icml/WangWLC24}       & \usym{1F5F8}  & \usym{2715} & \usym{2715} & \usym{2715}  & \usym{1F5F8} & \usym{2715} %&  \usym{1F5F8} 
        & $\mathcal{O}\big(\frac{\sigma^2}{\sqrt{TKm}}\big)+\mathcal{O}\big(\frac{\sqrt{K}\sigma_g^2}{\sqrt{Tm}}\big)  $$+\mathcal{O}\big(\frac{ \tau_{max}\tau_{avg_m} }{T}\big)$         \\ \hline
        TMC'25\cite{DBLP:journals/tmc/WangLJZ25}       & \usym{1F5F8}  & \usym{2715} & \usym{2715} & \usym{2715}  & \usym{2715} & \usym{2715} %&  \usym{1F5F8} 
        & $\mathcal{O}\big(\frac{\sigma^2+G^2}{\sqrt{Tn}}\big)$$+\mathcal{O}\big(\frac{\tau_{max}^2 }{T}\big)$
        \\ \hline
        {\bfseries AsynFL(this paper)}       &\usym{2714}  & \usym{2715} & \usym{2715} & \usym{2715}  & \usym{2714} & \usym{2714} %&  \usym{2714} 
        & $\mathcal{O}\big(\frac{\sigma^2}{\sqrt{TKn}}\big)  $$+\mathcal{O}\big(\frac{ \tau_{max}\tau_{avg_m} }{T}\big)$ ($n>m$)
            %\\ \hline
        %{\bfseries AsynFLC(this paper)}       &\usym{2714}  & \usym{2714} & \usym{2714} & \usym{2715}  & \usym{2714} & \usym{2714} & $\mathcal{O}\Big(\frac{ \tau_{max} \sigma^2 }{K \sqrt[3]{T^2}} \Big)$(IID)
            \\ \hline
        {\bfseries AsynFLC-EF  (this paper)}       &\usym{2714}  & \usym{2714} & \usym{2714} & \usym{2714}  & \usym{2714} & \usym{2714} %&  \usym{2714} 
        & $\mathcal{O}\big(\frac{\sigma^2}{\sqrt{TKn}}\big)  $$+\mathcal{O}\big(\frac{ \tau_{max}^2 }{T}\big)$   
		%\bottomrule % 底部线
	\\
        \hline  % 表格的横线
	\end{tabular*}
\label{table2}
\end{table*}

{\bfseries Comparisons with prior studies.} As summarized in TABLE \ref{table2}, we compare our convergence analysis with recent advances in asynchronous FL for non-convex optimization. Existing studies \cite{DBLP:conf/aistats/NguyenMZYR0H22,DBLP:conf/allerton/ToghaniU22,DBLP:conf/iclr/WangCWCC24,DBLP:conf/icml/WangWLC24,DBLP:journals/tmc/WangLJZ25,DBLP:journals/pami/LiYRSZ24} rely on a rather demanding assumption that all clients participate with uniform probability, which is often impractical. In contrast, our analysis eliminates this requirement, offering greater practical applicability. Furthermore, our framework imposes no constraints on the number of participating clients per round, enhancing flexibility and scalability. A third major advantage lies in the improved tightness of our convergence bound. For example, regarding the dominant term, our analysis achieves a tighter convergence rate of $\mathcal{O}\big(\frac{\triangle +\sigma^2}{\sqrt{TKn}}\big)$ preferable to $\mathcal{O}\big(\frac{\triangle +\sigma^2}{\sqrt{TKm}}\big)$$+\mathcal{O}\big(\frac{\sqrt{K}\sigma_g^2}{\sqrt{Tm}}\big)$ in the analyses \cite{DBLP:conf/iclr/WangCWCC24,DBLP:conf/icml/WangWLC24} where data heterogeneity $\sigma_g^2$ degrades the convergence at a rate of $\mathcal{O}\big(\frac{ 1}{\sqrt{T}}\big)$. For the non-dominant term, our bound of $\mathcal{O}\big(\frac{ \tau_{max}\tau_{avg_m} }{T}\big)$ matches that of \cite{DBLP:conf/iclr/WangCWCC24,DBLP:conf/icml/WangWLC24}. And this is superior to the bound $\mathcal{O}\big(\frac{\tau_{max}^2 }{T}\big)$ in other recent works.

{\bfseries Robustness against flexible participation, data heterogeneity and asynchronous delay.} Under flexible participation—where both the set of participating clients and their participation probabilities may vary non-uniformly per round, our analysis reveals that the dominant convergence term $\mathcal{O}\big(\frac{\triangle}{\sqrt{TKn}}+\frac{\sigma^2}{\sqrt{TKn}}\big)$ only depends on initialization and local stochastic variance$\sigma$, not being affected by data heterogeneity $\sigma_g$ and  asynchronous delay. The non-dominant term $\mathcal{O}\big(\frac{ \tau_{avg} \sigma^2 + Kn\tau_{max}\tau_{avg_m}  \sigma_g^2}{T}\big)$, decays at a faster rate of $\mathcal{O}\big(\frac{ 1}{T}\big)$, despite being jointly influenced by data heterogeneity $\sigma_g^2$ and the delay product $\tau_{max}\tau_{avg_m}$. These results demonstrate the robustness of AsynFL against flexible participation, data heterogeneity and asynchronous delay. 

{\bfseries The interaction between asynchronous delay and data heterogeneity.} This nonlinear coupling $\mathcal{O}\big(\frac{  \tau_{max}\tau_{avg_m}  \sigma_g^2}{T}\big)$ indicates a mutual exacerbation between asynchronous delay and data heterogeneity: each amplifies the adverse effect of the other on convergence. Consider the special case of IID data, i.e., $\sigma_g=0$, AsynFL converges at a rate of $\mathcal{O}\big(\frac{\triangle +\sigma^2}{\sqrt{TKn}}\big)$$ + \mathcal{O}\big(\frac{\tau_{avg} \sigma^2}{T}\big)$. The impact of asynchronous delay is limited to the average delay $\tau_{avg}$, which is significantly smaller than the product $\tau_{max} \tau_{avg_m}$. This indicates that in the absence of data heterogeneity, the coupling effect vanishes, and the influence of asynchronous delay is markedly reduced. When the number of communication rounds $T$ is sufficiently larger than $\Omega \big( Kn\tau_{avg}^2 \big) $, the impact of the delay becomes negligible. Therefore, convergence can be substantially accelerated by mitigating data heterogeneity or decreasing delays.

\subsection{Convergence Analysis of AsynFLC}

In the following, we analyze how biased compression interacts with asynchronous updates and non-IID data, and study how they jointly affect convergence for non-convex optimization. The proofs are provided in Appendix C of the supplementary material.

\begin{theorem}
\label{thm:2}
(Convergence of AsynFLC). Suppose Assumption \ref{ass:1} and Assumption \ref{ass:2} hold. If the local learning rates satisfy $\eta \leq  \frac{1}{4\sqrt{2-\gamma} (\tau_{max}+1)^{3/2} \tau_{max} \eta_g K L} $ and the relationship between the compression rate and asynchronous delay satisfies $1-\gamma \leq \frac{1}{2 (\tau_{max}+1)}$, AsynFLC satisfies:
\begin{equation}
\begin{aligned}
& \frac{1}{T}  \sum_{t=0}^{T-1}  \mathbb{E}\|\nabla f (\mathbf{x}_t)\|^2 \leq  \frac{4\big[ f(\mathbf{x}_0) - f (\mathbf{x}^{*}) \big]}{\eta_g \eta K T} \\ & +  4\big(\tau_{max}+1\big) \big(\tau_{avg}+1\big) \eta^2  K L^2 \sigma^2  + \\ & 16\big(\tau_{max}+1\big)  \tau_{avg}  \tau_{max} \big(  \eta^2 K L^2 + 2 \tau_{max} \eta^4 K^3 L^4\big) \sigma^2 \\ & +  16\big(\tau_{max}+1\big) \tau_{max}^3  \big(4  \eta^2 K^2 L^2 + 24  \eta^4 K^4 L^4\big) \sigma_g^2  \\ & +  48 \big(\tau_{max}+1\big)^2 \eta^2  K^2 L^2 \sigma_g^2  +  2 \big(\tau_{max}+1\big)^2 \sigma_g^2.
\end{aligned}
\end{equation}
\end{theorem}

By selecting a delay-dependent local learning rate and an appropriate global learning rate, we obtain the corollaries.

\begin{corollary}
\label{cor:2.0}
Suppose the conditions in Theorem \ref{thm:2} are satisfied. Let $\triangle =f(\mathbf{x}_0) - f (\mathbf{x}^{*})$. If data is IID, i.e., $\sigma_g=0$, and the relationship between compression rate and asynchronous delay satisfies  $1-\gamma \leq \frac{1}{2 (\tau_{max}+1)}$, choosing $\eta=\Theta\big(\frac{1}{ K \tau_{max} T^{1/3}}\big)$, $\eta_g=\Theta\big(K \big)$, AsynFLC satisfies:
\begin{equation}
\begin{aligned} 
\frac{1}{T} \sum_{t=0}^{T-1}  \mathbb{E}\|\nabla & f(\mathbf{x}_t)\|^2 =  \mathcal{O}\Big(\frac{ \tau_{max}(\triangle + \sigma^2) }{K \sqrt[3]{T^2}} \Big) .
\end{aligned}
\end{equation}
\end{corollary}

%\cref{thm:2} suggests that, the condition for the convergence of AsynFLC is that the compression coefficient satisfies $1-\gamma \leq \frac{1}{2 (\tau_{max}+1)}$ depending on the maximum delay. It is a demanding condition, indicating that only under extremely low compression rates AsynFLC may converge to a stationary point. 

\begin{remark}
    For convex optimization, counter-examples using Top$_k$ and Sign compressors have been provided to show the non-convergence issue of directly applying biased compression in distributed learning with IID data \cite{DBLP:conf/icml/KarimireddyRSJ19,DBLP:journals/jmlr/BeznosikovHRS23}.
\end{remark}

%\begin{remark}
    %Using only the upper bound provided by Theorem 4.4 \cite{DBLP:conf/icml/Li023} cannot prove the non-convergence issue of directly applying biased compression in non-convex synchronous FL. %The upper bound may be too loose. To prove the non-convergence issue, counterexamples or the lower bound that does not converge to zero are required.
%\end{remark}

%{\bfseries Sufficient conditions for the convergence of AsynFLC in non-convex optimization.} Non-convex problems may have multiple local optimal solutions. Corollary \ref{cor:2.0} shows the sufficient conditions for the convergence of AsynFLC to reach the optimum. If data is IID, i.e., $\sigma_g=0$, and the relationship between compression rate and asynchronous delay satisfies  $1-\gamma \leq \frac{1}{2 (\tau_{max}+1)}$, AsynFLC with biased compression can achieve a convergence rate of $\mathcal{O}\big(\frac{ \tau_{max}(\triangle + \sigma^2) }{K \sqrt[3]{T^2}}\big)$. 

{\bfseries Sufficient conditions for the convergence of AsynFLC in non-convex optimization.} Corollary \ref{cor:2.0} shows the sufficient conditions for the convergence of AsynFLC to reach the optimum. If data is IID, i.e., $\sigma_g=0$, and the relationship between compression rate and asynchronous delay satisfies  $1-\gamma \leq \frac{1}{2 (\tau_{max}+1)}$, AsynFLC with biased compression can achieve a convergence rate of $\mathcal{O}\big(\frac{ \tau_{max}(\triangle + \sigma^2) }{K \sqrt[3]{T^2}}\big)$.

{\bfseries Interaction between asynchronous delay and compression rate.} The constraint $1-\gamma \leq \frac{1}{2 (\tau_{max}+1)}$ captures a critical trade-off between the compression rate and asynchronous delay. Specifically, as the maximum delay $\tau_{max}$ increases, the allowable compression rate $(1-\gamma)$ must decrease to maintain optimization stability and ensure effective convergence. While higher compression rates are desirable to reduce communication overhead and improve efficiency, excessive compression under large asynchronous delays may lead to instability and poor convergence. This interaction enables adaptive strategies to optimize the trade-off between communication efficiency and convergence performance.
%This constraint indicates the balance required in practical implementations.
%This enables adaptive strategies that can dynamically adjust the compression rate in response to varying levels of asynchronous delay, thereby optimizing the trade-off between communication efficiency and convergence performance.
%depending on the maximum delay. It is a demanding condition, indicating that only under extremely low compression rates AsynFLC may converge to a stationary point. In the homogeneous setting where the local objective functions $f_i$ are identical, i.e., $\sigma_g=0$, AsynFLC with biased compression obtains a convergence rate of $\mathcal{O}\big(\frac{ \tau_{max}(\triangle + \sigma^2) }{K \sqrt[3]{T^2}}\big)$ only under extremely low compression rates.

\begin{corollary}
\label{cor:2}
Suppose the conditions in Theorem \ref{thm:2} are satisfied. Let $\triangle =f(\mathbf{x}_0) - f (\mathbf{x}^{*})$. If choosing $\eta=\Theta\big(\frac{1}{ K \tau_{max} T^{1/3}}\big)$, $\eta_g=\Theta\big(K \big)$, AsynFLC satisfies:
\begin{equation}
\begin{aligned}
\label{equ:15}
\frac{1}{T} \sum_{t=0}^{T-1}  \mathbb{E}\|\nabla & f(\mathbf{x}_t)\|^2 =  \mathcal{O}\Big(\frac{ \tau_{max}(\triangle + \sigma^2 ) }{K \sqrt[3]{T^2}} \\ & + \frac{ \tau_{max}^2 \sigma_g^2}{ \sqrt[3]{ T^2}} + \tau_{max}^2 \sigma_g^2\Big) .
\end{aligned}
\end{equation}
\end{corollary}

{\bfseries Impact of data heterogeneity on biased compression.} In non-IID settings where $\sigma_g^2>0$, the last term $\mathcal{O}\big(\tau_{max}^2 \sigma_g^2\big)$ in Equation \eqref{equ:15} does not decline. Data heterogeneity reflects divergence among clients' models, leading to more dramatic changes in both directions and magnitudes of local gradients. Such changes necessitate the retention of as much gradient information as possible to ensure accurate gradient estimation. However, when biased compression is applied to gradients, the lack of gradient information becomes severe, amplifying estimation variance. Consequently, in scenarios with high data heterogeneity, the direct application of biased compression can result in substantial degradation of model performance.

%The sign information alone is insufficient to accurately reflect the true gradient directions in such cases. Consequently, the value of $\gamma$ is reduced due to the increased heterogeneity of the data.

{\bfseries Joint impact of biased compression, asynchronous delay and data heterogeneity.} Compared to Theorem \ref{thm:1} for AsynFL without compression, the error part in Theorem \ref{thm:2} becomes larger due to the variance of gradient estimation caused by biased compression. Furthermore, the convergence bound depends on $\tau_{max}^4$, indicating the effect of asynchronous delay is exacerbated by biased compression. In summary, the non-vanishing term $\mathcal{O}\big(\tau_{max}^2 \sigma_g^2\big)$ in Corollary \ref{cor:2} stems indirectly from the large gradient variance caused by compression errors—an effect further exacerbated by both data heterogeneity and asynchronous delays. This indicates that AsynFLC only with biased compression is difficult to converge to a stationary point on non-IID data. %In summary, directly applying biased compression greatly impacts the convergence for asynchronous federated optimization.
%This demonstrates the impact of data heterogeneity and asynchronous delays on convergence.

\subsection{Convergence Analysis of AsynFLC-EF}

In the following,  we analyze how EF influences the convergence of AsynFLC-EF, especially under the joint impact of asynchronous delay and flexible participation. The proofs are provided in Appendix B of the supplementary material.
\begin{lemma}
\label{lemma}
(Bounded Accumulated Error). Suppose Assumption \ref{ass:1} and Assumption \ref{ass:2} hold. Let $c=\frac{ (1-\gamma)(2-\gamma)}{\gamma^2}$ denote the compression factor. If we choose $\eta=\Theta\big(\frac{1}{K \sqrt{T}}\big)$, $\eta_g=\Theta\big( \sqrt{Kn}\big)$, the average accumulated error $\mathbf{e}_{t}$ under $\ell_2$ norm in Algorithm \ref{alg:alg1} can be bounded:
\begin{equation}
\begin{aligned}  \frac{1}{T} \sum_{t=0}^{T-1}  \frac{1}{n} \sum_{i=1}^n \mathbb{E}\|\mathbf{e}_{t}^{(i)}\|^2  \leq  \mathcal{O}\Big\{c\cdot  \big( \frac{\tau_{max} \sigma^2}{KT}  + \frac{\tau_{max}^2 \sigma_g^2}{T} \big)\Big\}.  
\end{aligned}
\end{equation}
\end{lemma}

Lemma \ref{lemma} shows that the residual error converges rapidly to zero at a rate of $\mathcal{O}(\frac{1}{T})$. This result demonstrates that, through EF, 'important' errors with large magnitudes have been effectively compensated in the gradients, leaving only minor and negligible residual errors that diminish rapidly with increasing communication rounds $T$. The bound captures the combined effects of compression bias, asynchronous delay, and data heterogeneity, confirming the robustness and convergence of AsynFL-EF under biased compression, non-IID data and partial  participation.

\begin{theorem}
\label{thm:3}
(Convergence of AsynFLC-EF). Suppose Assumption \ref{ass:1} and Assumption \ref{ass:2} hold. If the local learning rates satisfy $\eta \leq \frac{\gamma}{72 \sqrt{3 (\gamma-1)^2 + 1} \tau_{max}^{1.5} \eta_g K L} $, AsynFLC-EF satisfies:
\begin{equation}
\begin{aligned} 
& \frac{1}{T}  \sum_{t=0}^{T-1} \mathbb{E}\|\nabla f (\mathbf{x}_t)\|^2 \leq \frac{8[ f(\mathbf{x}_0) - f (\mathbf{x}^{*}) ]}{\eta \eta_g K T}  +  \frac{4 \eta \eta_g  L\sigma^2}{n}  + \\ &  \eta^2 K L^2 \Big[ 4(\sigma^2+6 K \sigma_g^2\big)  + \frac{ 1 }{n} ( 42 \tau_{avg_0} + 144 \tau_{avg_1} )  \eta_g^2 \sigma^2 \Big]  \\ & + ( \lambda_1 + 2 \lambda_2 +  \lambda_3 )\eta^2  K L^2 \sigma^2   \\ & +  (\varphi_1 + 2 \varphi_2 + 2 \varphi_3 +  \varphi_4 )\eta^2  K^2 L^2 \sigma_g^2  ,
\end{aligned}
\end{equation}
where $\lambda_1$, $\lambda_2$, $\varphi_1$, $\varphi_1$, $\varphi_2$ and $\varphi_3$ can be found in Theorem \ref{thm:1}. Besides, $\lambda_3 = \frac{108 (1-\gamma)(2-\gamma)}{ \gamma^2}( 3\tau_{max}  \eta_g^2  + 6\tau_{max}^2  \eta^2 \eta_g^2 K^2 L^2 ) $, $\varphi_4 =\frac{108 (1-\gamma)(2-\gamma)}{ \gamma^2} \tau_{max}^2 (12  \eta_g^2  + 72  \eta^2 \eta_g^2 K^2 L^2 ) $.
\end{theorem}

\begin{corollary}
\label{cor:3}
Suppose the conditions in Theorem \ref{thm:3} are satisfied. Let $\triangle =f(\mathbf{x}_0) - f (\mathbf{x}^{*})$, choosing $\eta=\Theta\big(\frac{1}{K \sqrt{T}}\big)$, $\eta_g=\Theta\big( \sqrt{Kn}\big)$, AsynFLC-EF satisfies:
\begin{equation}
\begin{aligned}
\frac{1}{T} \sum_{t=0}^{T-1}  \mathbb{E}&\|\nabla f(\mathbf{x}_t)\|^2 =  \mathcal{O}\Big(\frac{\triangle}{\sqrt{TKn}}+\frac{\sigma^2}{\sqrt{TKn}} 
 +  \frac{\sigma^2+K \sigma_g^2 }{TK} \\ & + \frac{ (1-\gamma)(2-\gamma)}{ \gamma^2} \frac{ n\tau_{max} \sigma^2 + Kn\tau_{max}^2 \sigma_g^2}{T} \Big) .
\end{aligned}
\end{equation}
\end{corollary}

{\bfseries The effect of EF.} Theorem \ref{thm:3} provides a convergence upper bound that holds for any compression ratio. In contrast to Theorem \ref{thm:2} for AsynFLC without EF, the error part in Theorem \ref{thm:3} is smaller and can converge to zero. Compared to Theorem \ref{thm:1} for AsynFL without gradient compression, the error part in Theorem \ref{thm:3} has a similar convergence rate. Corollary \ref{cor:3} suggests that for a sufficiently large $T$, AsynFLC-EF achieves a convergence rate of $\mathcal{O}\big(\frac{1}{\sqrt{TKn}}\big)$ and a communication round complexity of $\mathcal{O}\big(\frac{1}{Kn\epsilon^2}\big)$ when reaching a $\epsilon-$stationary point, i.e., $\frac{1}{T}  \sum_{t=0}^{T-1} \mathbb{E}\|\nabla f(\mathbf{x}_t)\|^2 \leq \epsilon$. This indicates that, AsynFLC-EF has the same convergence rate and communication round complexity as AsynFL, but requires smaller communication costs. Compared to other full-precision asynchronous FL methods, AsynFLC-EF achieves the same convergence of $\mathcal{O}\big(\frac{1}{\sqrt{T}}\big)$$+\mathcal{O}\big(\frac{\tau_{max}^2}{T}\big)$ as in \cite{DBLP:conf/aistats/NguyenMZYR0H22,DBLP:conf/allerton/ToghaniU22}, but is more communication-efficient. These results indicate that, despite asynchronous updates, EF can effectively reduce the variance caused by biased gradient compression, which accounts for the great improvement in the convergence of AsynFLC-EF.

Furthermore, only the last term is affected by the compression rate and delay, which decays at a faster rate of $\mathcal{O}\big(\frac{1}{T}\big)$. When $T$ is sufficiently large, the impact of gradient compression and delay becomes negligible. This indicates that asynchronous FL with EF is robust against compressed and delayed gradients. This is important for asynchronous federated optimization to reduce communication costs.

{\bfseries The impact of asynchronous delay and flexible participation on EF.} Compared to Theorem \ref{thm:1} for AsynFL without gradient compression, the error part in Theorem \ref{thm:3} becomes slightly larger due to the additional terms where $\lambda_3 $ and $\varphi_4$ are related to the compression rate. Corollary \ref{cor:3} also shows that the higher order convergence term relies on compression rate $\frac{ (1-\gamma)(2-\gamma)}{ \gamma^2}$ and the maximum delay $\tau_{max}^2$. It can be explained from two aspects. First, under EF, compression errors are compressed with subsequent updates, and smaller gradient contributions are omitted, resulting in a slowdown factor of $\frac{ (1-\gamma)(2-\gamma)}{ \gamma^2}$. Second, in asynchronous settings—especially under flexible client participation—error compensation is delayed until the client’s next participation. Compared to Corollary \ref{cor:1} for AsynFL, the convergence of AsynFLC-EF greatly depends on the maximum delay $\tau_{max}^2$. The combined impact on EF is a compound slowdown of $\frac{ (1-\gamma)(2-\gamma)}{ \gamma^2}\cdot \tau_{max}^2$. This multiplicative interaction explains why EF cannot restore the compressed gradients to full precision. Flexible participation changes disrupt training continuity, while unpredictable error compensation timing exacerbates delays. This prevents timely correction of locally accumulated compression errors, causing progressive increases in global gradient variance and biased model updates that ultimately hinder convergence.

%biased gradient compression allows only partial gradient update information to be sent to the server, resulting in significant deviation and variance.
%Compared to other full-precision asynchronous FL methods, AsynFLC-EF achieves the same convergence of $\mathcal{O}\big(\frac{1}{\sqrt{T}}\big)$$+\mathcal{O}\big(\frac{\tau_{max}^2}{T}\big)$ as in \cite{DBLP:conf/aistats/NguyenMZYR0H22,DBLP:conf/allerton/ToghaniU22}, but is more communication-efficient.

\section{Experiments}
To validate the correctness of our theoretical analysis, we conduct an extensive set of simulation experiments. Specifically, we evaluate the convergence behaviors of different AFL methods, examine the efficiency of various compression strategies, and assess the impact of client participation and data heterogeneity.  

\subsection{Experimental Settings}

\textbf{Datasets and models.} We conduct experiments on three popular datasets: (a) MNIST; (b) FMNIST; (c) CIFAR-10. For MNIST and FMNIST, we train the typical MLP model with a local learning rate of 0.01. For CIFAR-10, we train three popular models, i.e., CNN with a local learning rate of 0.01, AlexNet with a local learning rate of 0.0001.

\textbf{Training setup.} We test \textit{n} = 100 clients. For this experiment, we generate non-IID local data using a Dirichlet distribution with parameter 0.4, the same approach as in~\cite{DBLP:conf/aistats/NguyenMZYR0H22}. The local mini-batch size is 128.  In addition, to simulate the asynchronism, we assume that the training time of a client follows a normal distribution.
%We run \textit{T} = 3000 rounds, where one FL training round is started after server waits for the specified time.

\textbf{Methods and compressors.} We compare the following FL training methods/algorithms in our experiments:
    \begin{itemize}
    \item \textbf{FedBuff:} an asynchronous FL framework where clients upload their local updates with full precision and the server caches a specified number of these local updates for the global update \cite{DBLP:conf/aistats/NguyenMZYR0H22}. 
    
    \item \textbf{AsynFL:} an asynchronous FL framework which enables the number of  participating clients to flexibly vary in each round. The framework does not apply gradient compression techniques.

    \item \textbf{AsynFLC(signSGD):} AsynFLC that directly applies SignSGD compressor~\cite{DBLP:conf/icml/KarimireddyRSJ19} \textbf{without EF}.
    
    \item \textbf{AsynFLC(topk):} AsynFLC that directly applies Top$_k$ compressor~\cite{DBLP:conf/nips/AlistarhH0KKR18} \textbf{without EF}. For this compressor, we test parameter \textit{k/d} $\in$ \{0.03, 0.06, 0.1\}.

    \item \textbf{AsynFLC(QSGD):} AsynFLC that applies unbiased quantization (QSGD) compressor~\cite{DBLP:conf/nips/AlistarhG0TV17} \textbf{without EF}. For this compressor, we test parameter \textit{b} $\in$ \{2, 4, 8\}.

    \item \textbf{AsynFLC-EF(topk):} AsynFLC-EF with Top$_k$ compressor and \textbf{EF}.

    \item \textbf{AsynFLC-EF(topk+QSGD):} AsynFLC-EF with the combination of QSGD and Top$_k$ compressor and EF (i.e., a further compression over Top$_k$ under same sparsity). For this compressor, we test parameter \textit{k/d} $\in$ \{0.03, 0.06, 0.1\} and \textit{b} $\in$ \{2, 4, 8\}.
    \end{itemize}

\subsection{Experimental Results}

% \begin{table}[!htbp]
% \renewcommand{\arraystretch}{1.6}  % 增大行高
% \setlength{\tabcolsep}{3pt}  % 调整列间距
% \small
% \caption{Compression ratio of different methods compared to full precision communication.}
% \centering
% \begin{tabular}{|c|c|}
% \hline
%     Method   &    Compression ratio (\%) \\ \hline
%     FedBuff   & 100 \\
%     AsynFL    & 100 \\
%     AsynFLC(Q4b)   & 12.5 \\
%     AsynFLC-EF(top10)   & 10 \\
%     AsynFLC-EF(top6)   & 6 \\
%     AsynFLC-EF(top3)   & 3 \\
%     AsynFLC-EF(top10+Q4b)   & 1.25 \\
%     AsynFLC-EF(top6+Q4b)   & 0.75 \\
%     AsynFLC-EF(top3+Q8b)   & 0.75 \\
%     AsynFLC-EF(top3+Q4b)   & 0.375 \\
%     AsynFLC-EF(top3+Q2b)   & 0.1875 \\
% \hline
% \end{tabular}
% \end{table}

\begin{figure*}
\centering
\includegraphics[width=1\textwidth]{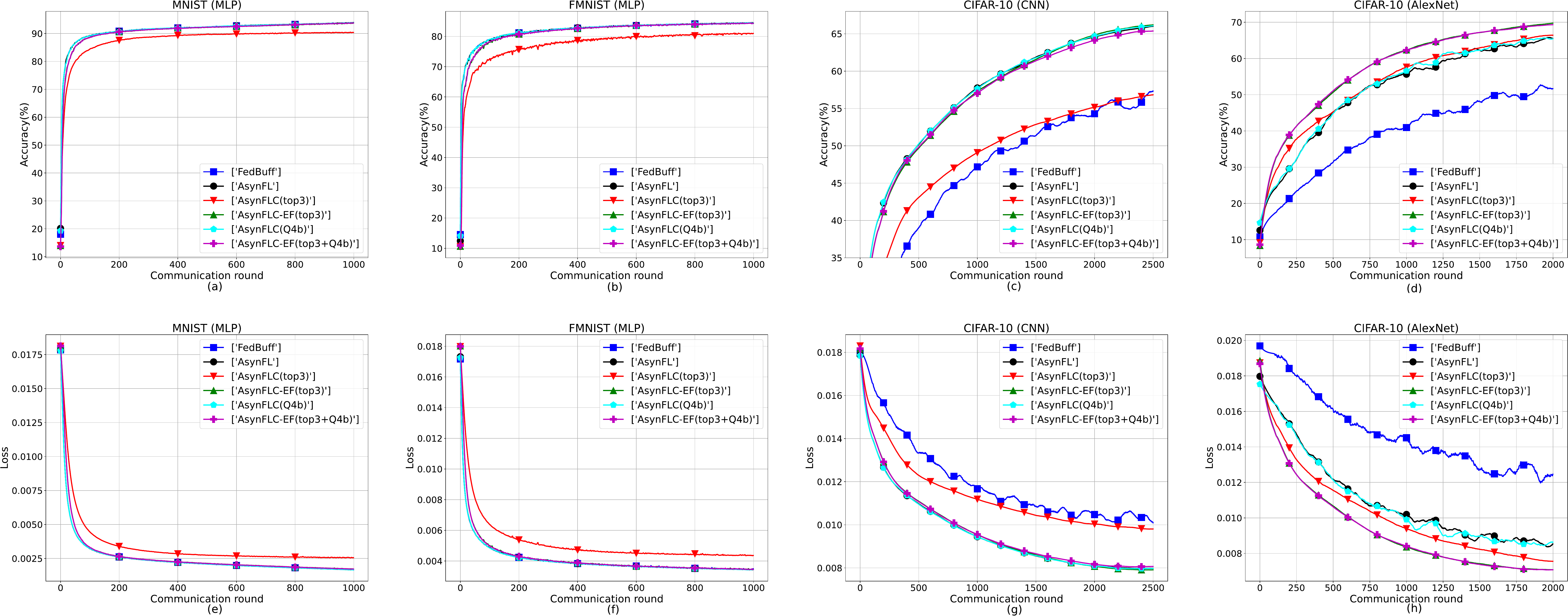}
\caption{Comparison of the accuracy and loss of FedBuff, AsynFL, AsynFLC(top3), AsynFLC-EF(top3), AsynFLC(Q4b), and AsynFLC-EF(top3+Q4b) on MNIST(MLP), FMNIST(MLP), CIFAR-10(CNN), and CIFAR-10(AlexNet).}
\label{fig1}
\end{figure*}

\begin{figure*}[ht]
\centering
\includegraphics[width=1\textwidth]{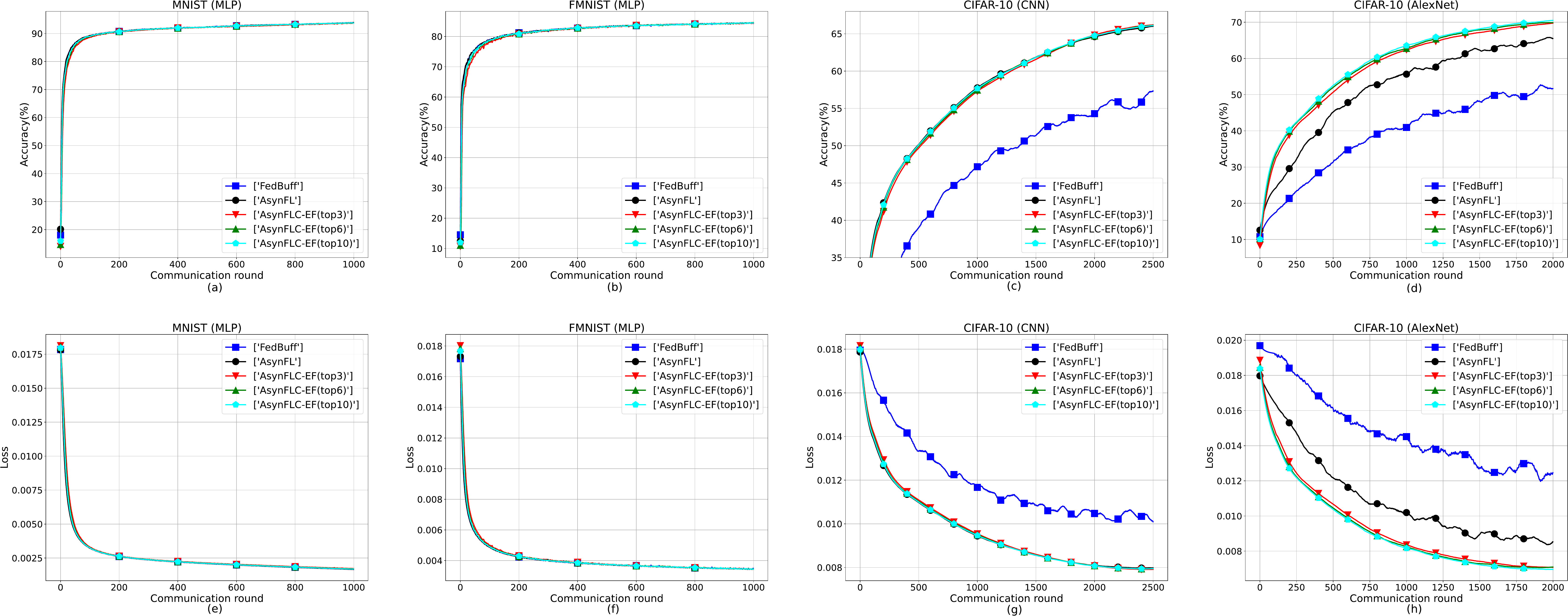}
\caption{Comparison of the accuracy and loss of FedBuff, AsynFL, AsynFLC-EF(top3), AsynFLC-EF(top6), and AsynFLC-EF(top10) on MNIST(MLP), FMNIST(MLP), CIFAR-10(CNN), and CIFAR-10(AlexNet).}
\label{fig2}
\end{figure*}

\begin{figure*}[ht]
\centering
\includegraphics[width=1\textwidth]{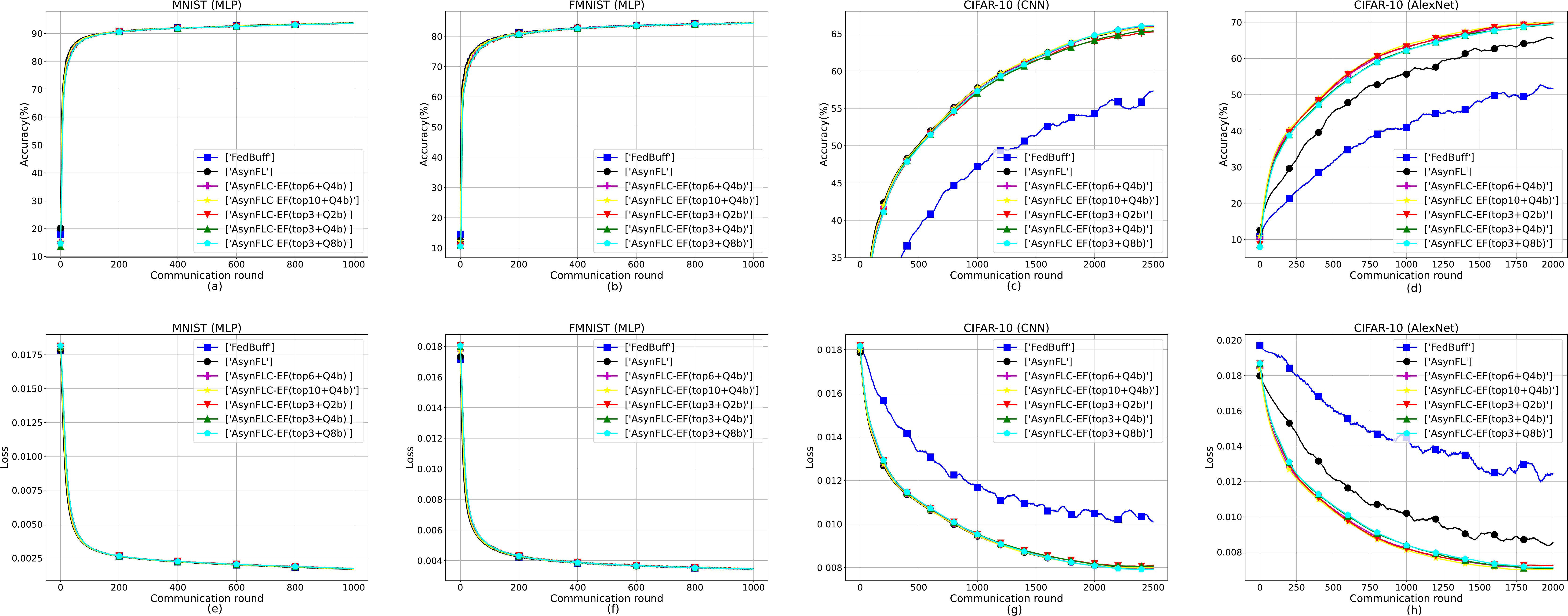}
\caption{Comparison of the accuracy and loss of FedBuff, AsynFL, AsynFLC-EF(top6+Q4b), AsynFLC-EF(top10+Q4b), AsynFLC-EF(top3+Q2b), AsynFLC-EF(top3+Q4b), and AsynFLC-EF(top3+Q8b) on MNIST(MLP), FMNIST(MLP), CIFAR-10(CNN), and CIFAR-10(AlexNet).}
\label{fig3}
\end{figure*}

\begin{figure*}[ht]
\centering
\includegraphics[width=1\textwidth]{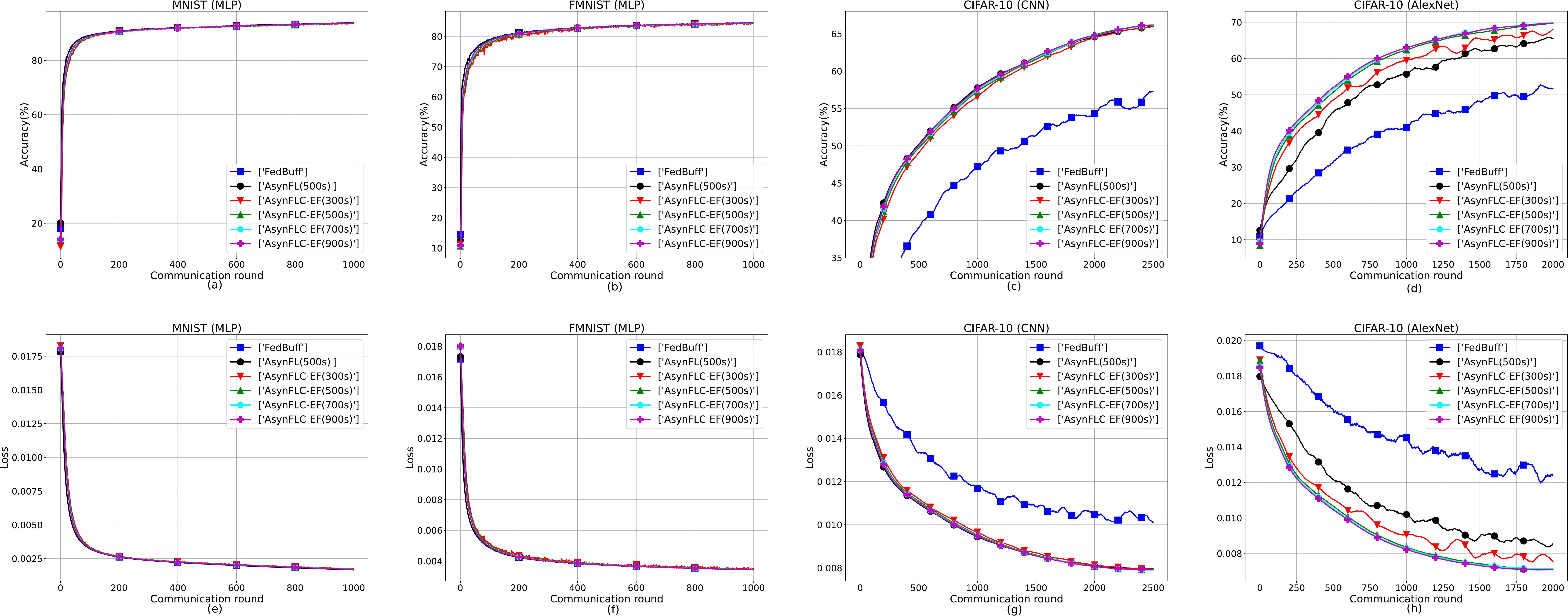}
\caption{Comparison of the accuracy and loss of FedBuff, AsynFL(500s), AsynFLC-EF(300s), AsynFLC-EF(500s), AsynFLC-EF(700s), and AsynFLC-EF(900s) on MNIST(MLP), FMNIST(MLP), CIFAR-10(CNN), and CIFAR-10(AlexNet).}
\label{fig4}
\end{figure*}

\begin{figure*}[ht]
\centering
\includegraphics[width=1\textwidth]{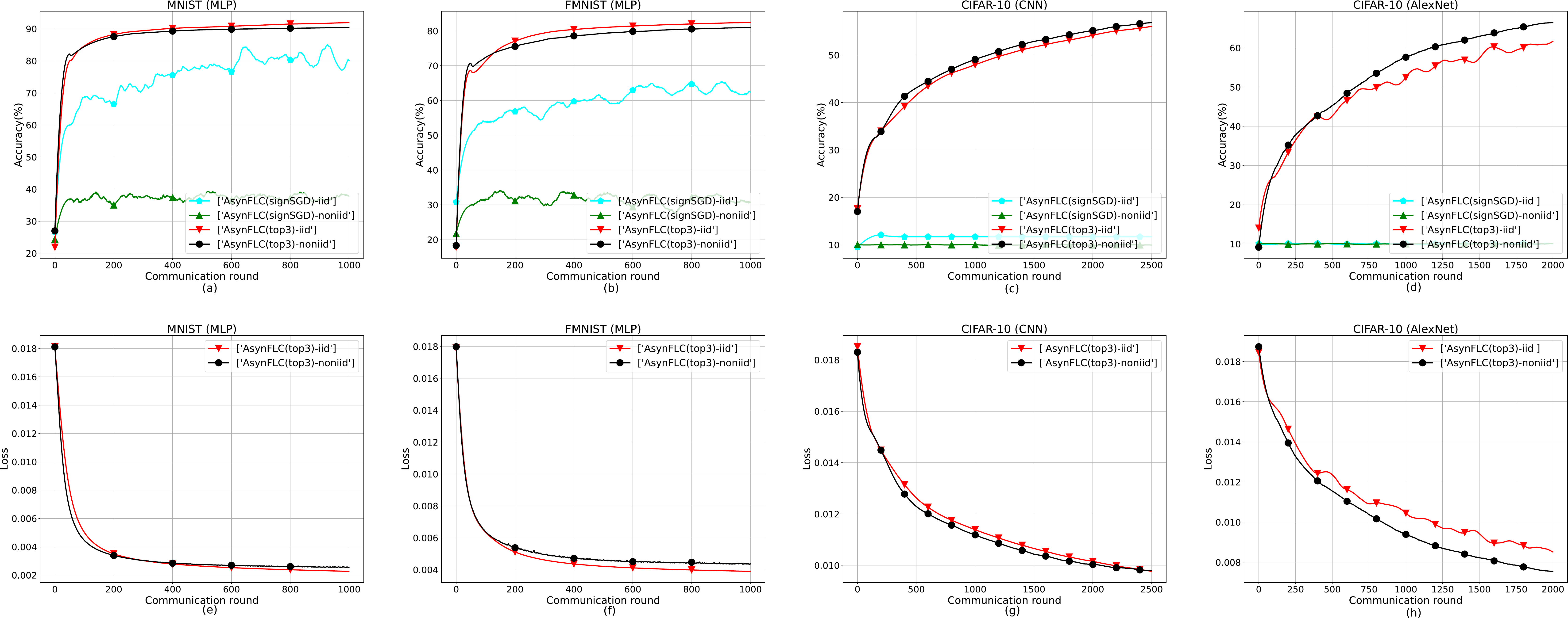}
\caption{Comparison of the accuracy and loss of AsynFLC(signSGD)-iid/noniid, AsynFLC(top3)-iid/noniid. (a) MLP trained on MNIST, (b) MLP trained on FMNIST, (c) CNN trained on CIFAR-10, and (d) AlexNet trained on CIFAR-10.}
\label{fig5}
\end{figure*}

\begin{table*}[tbp]
\renewcommand{\arraystretch}{1.9}  % 增大行高
\setlength{\tabcolsep}{2pt}  % 调整列间距
\small
\centering
\caption{The communication cost (in GB) required to achieve the specified accuracy on the specified dataset.}
\scalebox{0.92}{
\begin{tabular}{c|>{\centering\arraybackslash}m{2cm}>{\centering\arraybackslash}m{2.5cm}>{\centering\arraybackslash}m{2.5cm}>{\centering\arraybackslash}m{2.5cm}>{\centering\arraybackslash}m{2.5cm}>{\centering\arraybackslash}m{2.5cm}}
\hline
        \diagbox{\makecell{dataset/model}}{method} & FedBuff       & AsynFL                & AsynFLC(Q4b)         & AsynFLC-EF(top10)    & AsynFLC-EF(top6)     & AsynFLC-EF(top3)     \\ \hline
\makecell{MNIST\\(MLP, 85\%)}   & 0.41       & 0.39                & 0.05                & 0.05                & 0.04                 & 0.02               \\ \hline
\makecell{FMNIST\\(MLP, 75\%)}   & 0.48       & 0.48                & 0.06                & 0.05                & 0.06                 & 0.02               \\ \hline
\makecell{CNN\\(CIFAR-10, 55\%)}   & 497.67        & 364.43                & 45.38                & 36.67                & 22.38                & 11.43               \\ \hline
\makecell{AlexNet\\(CIFAR-10, 55\%)} & 1078.75       & 765.66                & 92.08                & 47.94                & 29.69                & 15.71                \\ \hline
       \diagbox{dataset/model}{method} & AsynFLC(top3) & AsynFLC-EF(top10+Q4b) & AsynFLC-EF(top6+Q4b) & AsynFLC-EF(top3+Q8b) & AsynFLC-EF(top3+Q4b) & AsynFLC-EF(top3+Q2b) \\ \hline
\makecell{MNIST\\(MLP, 85\%)}   & 0.03       & 0.006                & 0.004                & 0.004                & 0.002                 & 0.001               \\ \hline
\makecell{FMNIST\\(MLP, 75\%)}   & 0.05       & 0.007                & 0.004                & 0.005                & 0.002                 & 0.001               \\ \hline
\makecell{CNN\\(CIFAR-10, 55\%)}    & 26.75         & 4.57                  & 2.77                 & 2.84                 & 1.41                 & 0.72                 \\ \hline
\makecell{AlexNet\\(CIFAR-10, 55\%)} & 15.40         & 5.88                  & 3.68                 & 3.95                 & 1.96                 & 0.89                 \\ \hline
\end{tabular}
}
\label{table3}
\end{table*}

\textbf{(1) Superior Performance of AsynFLC-EF: Achieving Faster Convergence and Lower Communication Costs}

From Fig.\ref{fig1} to Fig.\ref{fig4} and Table \ref{table3}, we compare the convergence behavior and communication costs of full-precision FedBuff with different asynchronous FL methods—AsynFL, AsynFLC, and AsynFLC-EF. Based on the experimental results, the following observations can be made:

1) AsynFL and AsynFLC-EF achieve faster convergence rates and lower communication costs compared to FedBuff, which aligns with the theoretical analyses provided in Theorem \ref{thm:1} and Theorem \ref{thm:3}. These advantages become even more pronounced as models and data become more complex. For instance, on CIFAR-10 using CNN and AlexNet, both AsynFL and AsynFLC-EF exhibit significantly accelerated convergence while substantially reducing communication overhead compared to FedBuff. As illustrated in Fig.\ref{fig2} to Fig.\ref{fig4}, this consistent advantage holds across various settings, including different compression methods and asynchronous settings. Moreover, as quantified in TABLE \ref{table3}, AsynFL and AsynFLC-EF require considerably lower communication costs.  From TABLE \ref{table3}, we can observe that AsynFLC-EF, combined with Top$_k$ and QSGD ( $k/d=0.03, b=2$), can reduce the communication cost to 0.89 GB, while Fedbuff requires 1078.75 GB on CIFAR-10 using AlexNet. Additionally, the communication costs demanded by FedBuff amount to 1.4 times those of AsynFL on CIFAR-10 using CNN/AlexNet when reaching the specified accuracy of 55$\%$. This suggests that AsynFL achieves better communication efficiency than FedBuff, which may be attributed to the flexible participation enabled by AsynFL. These results indicate the effectiveness of AsynFL and AsynFLC-EF: faster convergence, better accuracy, and lower communication costs.

Fig.\ref{fig1} also shows the convergence results of AsynFL, AsynFLC, and AsynFLC-EF. Major observations are as follows: 

2) AsynFLC-EF achieves slightly higher convergence rates and higher communication efficiency than unbiased QSGD. Compared to AsynFLC with unbiased QSGD,  AsynFLC-EF achieves faster convergence rates on CIFAR-10 with AlexNet. From TABLE \ref{table3}, we can observe that AsynFLC-EF with Top $3\%$ reaches 48.74$\times$, and AsynFLC-EF with the combination of Top $3\%$ and QSGD ($b=2$) reaches a compression ratio of 860.29$\times$ (communication costs under full precision / communication costs under compression), while AsynFLC with QSGD ($b=4$) achieves 8.25$\times$ with the same accuracy on CIFAR-10 using AlexNet. 

3) AsynFLC achieves worse convergence performance than AsynFL under non-IID data, due to the lack of EF. AsynFLC with Top $3\%$ shows unstable convergence compared to AsynFL and AsynFLC-EF, which results from the biased compression of gradients without EF. This is consistent with our analytical findings.

4) AsynFLC-EF achieves a convergence rate comparable to that of AsynFL, suggesting that the EF effectively counteracts the adverse effects of compression.  This can be explained by Theorem \ref{thm:2} and Theorem \ref{thm:3}, demonstrating that the biased compression causes a large variance and EF applied in asynchronous updates can still effectively reduce the variance, thereby maintaining the convergence speed.

%Such results also verify our analytical finding that AsynFLC-EF achieves nearly the same convergence as AsynFL. AsynFLC-EF is more efficient in communication as Top$_k$ sparsification can reduce more communication costs than quantization.

\textbf{(2) AsynFLC-EF Under Various Compression Strategies: Unifying Efficiency and Robustness}

Fig.\ref{fig2} and Fig.\ref{fig3} present the convergence performance of AsynFLC-EF under various compression strategies. %For Top$_k$ compressor, we test parameter \textit{k/d} $\in$ \{0.03, 0.06, 0.1\}. $k/d=0.03$ means that only 3 $\%$ parameters are uploaded to the server. For the combination of Top$_k$ and QSGD compressor, we test parameter $k/d=0.03$ with \textit{b} $\in$ \{2, 4, 8\} and \textit{k/d} $\in$ \{0.03, 0.06, 0.1\} with $b=4$.
%The following observations can be made:

5) AsynFLC-EF across various compression strategies can achieve convergence rates comparable to those of AsynFL, demonstrating its robustness—a finding consistent with the theoretical analysis provided in Theorem \ref{thm:3}. Furthermore, AsynFLC-EF achieves better accuracy with a substantial reduction in communication costs compared to AsynFL. For instance, TABLE \ref{table3} shows that AsynFLC-EF with the combination of Top $3\%$ and QSGD ($b=2$) achieves an 860$\times$ compression ratio when reaching the target accuracy on CIFAR-10 using the AlexNet model. It demonstrates the effectiveness of EF in the asynchronous and heterogeneous setting, improving communication efficiency without sacrificing model accuracy. 

6) AsynFLC-EF demonstrates robustness against compressed, asynchronous and heterogeneous gradient updates. As illustrated in Fig. \ref{fig2} and Fig. \ref{fig3}, even under non-IID data, partial participation, and asynchronous training environments, higher compression ratios do not compromise convergence speed or model accuracy. As shown in TABLE \ref{table3}, AsynFLC-EF with the combination of Top $3\%$ and QSGD ($b=2$) maintains rapid convergence and superior accuracy when reaching 860$\times$ compression ratio. 

\textbf{(3) Client Participation and Data Heterogeneity: Key Factors in FL Convergence}

Fig.\ref{fig4} illustrates the impact of varying waiting time values on the convergence rate of AsynFLC-EF. Different waiting times—specifically, 300, 500, 700, and 900 seconds—result in varying numbers of local updates received by the server, reflecting the flexible participation of clients. And 1000 means synchronous updates. The key observations are as follows: 

7) For CIFAR-10 using the AlexNet model, AsynFLC-EF attains a faster convergence speed when the server adopts a longer waiting time. For CIFAR-10 using the CNN model, AsynFLC-EF achieves very similar convergence rates and accuracy across different waiting times. 
These results indicate  that AsynFLC-EF remains robust under varying client participation patterns. Consequently, the waiting time can be adaptively configured in each training round to achieve faster convergence and higher efficiency. 

Fig.\ref{fig5} shows the impact of data heterogeneity on the convergence rate of AsynFLC. We compare the convergence performance of AsynFLC(signSGD), AsynFLC(top3) in the IID or non-IID case. The key observations are as follows: % (a) MLP trained on MNIST, (b) MLP trained on FMNIST, (c) CNN trained on CIFAR-10, (d) AlexNet trained on CIFAR-10, (e) VGG11 trained on CIFAR-10.  

8) Data heterogeneity hampers the convergence of AsynFLC without EF. As observed, under non-IID data, AsynFLC using the Sign compressor fails to converge, and AsynFLC with Top$_k$ on CIFAR-10 using AlexNet also fails to converge.

9) In the non-convex setting, with IID data, AsynFLC can attain a satisfactory convergence rate when employing Top$_k$ and Sign, provided that the compression rate is sufficiently low or the asynchronous delay is relatively small, without the need for EF. This can be demonstrated by Corollary \ref{cor:2.0}. When the Sign compressor is applied to larger models including CNN and AlexNet, AsynFLC fails to converge. This is attributed to the fact that as the dimension of the gradients increases, the absence of gradient magnitude information results in a higher compression error. This observation aligns with the analysis presented in Theorem \ref{thm:2}.

\section{Conclusion}
In this paper, we study biased gradient compression and EF in asynchronous FL. We conduct a comprehensive analysis of their interactions and combined impact on convergence. We prove that EF effectively helps AsynFLC-EF achieve the same convergence rate as the full-precision counterpart. Furthermore, we analyze the joint impact of non-IID data, asynchronous dalay, and flexible participation on the convergence of AsynFLC-EF. To the best of our knowledge, this should be the first study on the convergence of Asynchronous FL with gradient compression and EF.

\bibliographystyle{IEEEtran}
\bibliography{ref.bib}

\newpage

\vfill

\end{document}